\documentclass[10pt,twocolumn,letterpaper]{article}

\usepackage{iccv}
\usepackage{times}
\usepackage{epsfig}
\usepackage{graphicx}
\usepackage{amsmath}
\usepackage{amssymb}
\usepackage{graphicx}
\usepackage{amsmath}
\usepackage{amssymb}
\usepackage{booktabs}
\usepackage{multirow}
\usepackage{subcaption}
\usepackage{listings}
\usepackage{amsthm}
\usepackage{colortbl}
\usepackage{braket}
\usepackage{array}
\usepackage{float}
\usepackage{enumitem}
\usepackage{algorithm}
\usepackage{bbding}
\usepackage{tabu}
\usepackage{graphicx}
\usepackage{microtype}      
\usepackage[dvipsnames]{xcolor}
\usepackage{caption}
\usepackage{xspace}
\usepackage{comment}
\usepackage{tabulary,overpic}

\usepackage[pagebackref=true,breaklinks=true,letterpaper=true,colorlinks,bookmarks=false]{hyperref}

\usepackage{amsmath,amsfonts,bm}

\def\eqref#1{equation~\ref{#1}}

\def\1{\bm{1}}

\DeclareMathAlphabet{\mathsfit}{\encodingdefault}{\sfdefault}{m}{sl}
\SetMathAlphabet{\mathsfit}{bold}{\encodingdefault}{\sfdefault}{bx}{n}

\usepackage{url}
\def\red#1{\textcolor[rgb]{1,0,0}{#1}}

\usepackage[capitalize]{cleveref}
\crefname{section}{Sec.}{Secs.}
\Crefname{section}{Section}{Sections}
\Crefname{table}{Table}{Tables}
\crefname{table}{Tab.}{Tabs.}

\iccvfinalcopy 

\ificcvfinal\fi

\newcommand{\sixth}{6$^{\mathrm{th}}$\xspace}
\newcommand{\eighth}{8$^{\mathrm{th}}$\xspace}
\newcommand{\tenth}{10$^{\mathrm{th}}$\xspace}

\newcommand{\methodName}{DeepMIM\xspace}

\begin{document}

\title{\methodName: Deep Supervision for Masked Image Modeling}

\author{
Sucheng Ren$^{1}$\thanks{Equal contribution.} \quad Fangyun Wei$^{1}$\footnotemark[1]~~\thanks{Corresponding author: fawe@microsoft.com.} \quad Samuel Albanie$^{2}$ \quad Zheng Zhang$^{1}$ \quad Han Hu$^{1}$ \\
$^1$Microsoft Research Asia \quad $^2$University of Cambridge \\
}

\maketitle
\ificcvfinal\thispagestyle{empty}\fi

\begin{abstract}
Deep supervision, which involves extra supervisions to the intermediate features of a neural network, was widely used in image classification in the early deep learning era since it significantly reduces the training difficulty and eases the optimization like avoiding gradient vanish over the vanilla training. Nevertheless, with the emergence of normalization techniques and residual connection, deep supervision in image classification was gradually phased out. In this paper, we revisit deep supervision for masked image modeling (MIM) that pre-trains a Vision Transformer (ViT) via a mask-and-predict scheme. Experimentally, we find that deep supervision drives the shallower layers to learn more meaningful representations, accelerates model convergence, and expands attention diversities. Our approach, called \methodName, significantly boosts the representation capability of each layer. In addition, \methodName is compatible with many MIM models across a range of reconstruction targets. For instance, using ViT-B, \methodName on MAE achieves 84.2 top-1 accuracy on ImageNet, outperforming MAE by +0.6.
By combining \methodName with a stronger tokenizer CLIP, our model achieves state-of-the-art performance on various downstream tasks, including image classification (85.6 top-1 accuracy on ImageNet-1K, outperforming MAE-CLIP by +0.8), object detection (52.8 AP$^{\text{box}}$ on COCO) and semantic segmentation (53.1 mIoU on ADE20K). Code and models are available at \url{https://github.com/OliverRensu/DeepMIM}.

\if
Masked image modeling (MIM), which couples the mask-and-predict task with an encoder-decoder architecture, shows considerable potential for ViT pre-training.
A range of prior work has investigated refinements to MIM relating to the design of appropriate reconstruction targets.
In this work, we study an orthogonal dimension of MIM---where to apply the reconstruction loss.
We discover that features produced by shallower Transformer blocks exhibit predictive power for reconstruction, motivating the application of deep supervision on intermediate features during pre-training.
Rather than using the same reconstruction targets as the last block, we generate different levels of hybrid targets, that are less challenging to reconstruct as targets for less-discriminative intermediate features.
Our approach, called \methodName, significantly boosts the representation capability of each block.
In addition, \methodName is compatible with many MIM models across a range of architectures and reconstruction targets.
For instance, using ViT-B, \methodName on MAE achieves 84.2 top-1 accuracy on ImageNet, outperforming MAE by +0.6 while reducing training epochs by half.
By combining \methodName with a stronger tokenizer CLIP, our model achieves state-of-the-art performance on various downstream tasks, including image classification (85.6 top-1 accuracy on ImageNet-1K outperforms by MAE-CLIP by 0.8\%), object detection (52.8 AP$^{\text{box}}$ on COCO) and semantic segmentation (53.1 mIoU on ADE20K).
\fi
\end{abstract}

\section{Introduction}
\begin{figure}
\centering
    \includegraphics[width=1.0\linewidth]{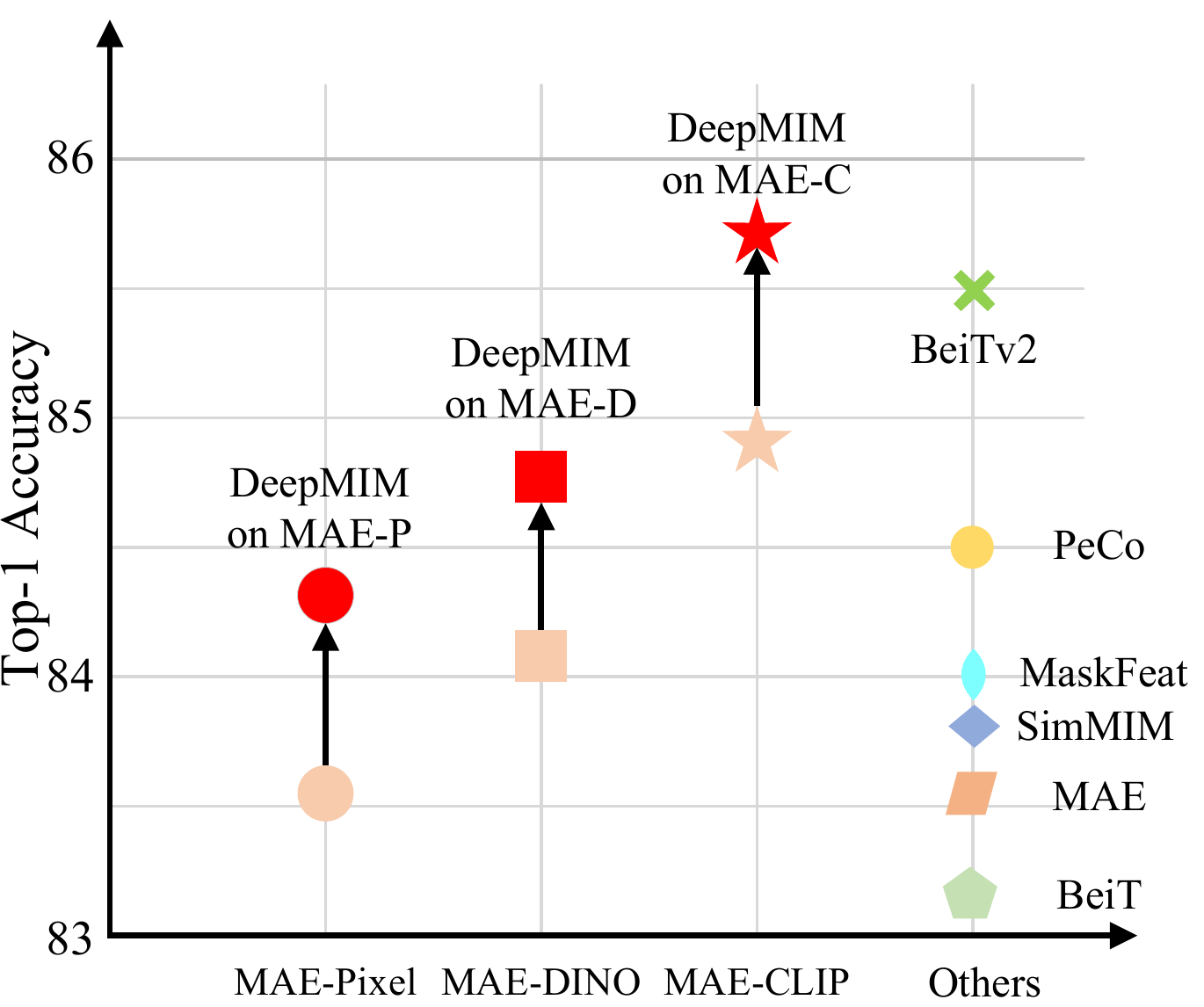}
    \caption{\methodName improves a broad range of MIM models.
    We highlight the improvements of top-1 accuracy on ImageNet-1K achieved by combining \methodName with various reconstruction targets (\textit{e.g.}, MAE-Pixels, MAE-DINO and MAE-CLIP). All models are pre-trained for 1600 epochs except those shown on the far right which are the best-performing models under the 800-epoch or 1600-epoch pre-training schedule.}
    
    \label{fig:tesear_acc}
\end{figure}
In the early deep learning era, deep supervision, which involves extra supervisions to the intermediate features of a neural network, has been widely adopted to a number of computer vision tasks, e.g., image classification~\cite{szegedy2015going,lee2015deeply} and edge detection~\cite{xie2015holistically}. Almost a decade ago, GoogLeNet~\cite{szegedy2015going} won the ILSVRC14 competition. A key element of GoogLeNet was its use of additional losses on intermediate features, motivated by the goal of ensuring gradients are propagated back through the network effectively, and to improve regularisation and discrimination in the network lower layers. However, deep supervision in image classification has fallen out of favour with the emergence of batch normalization~\cite{ioffe2015batch} and residual connections~\cite{he2016deep} which appear to have substantially mitigated issues relating to vanishing gradients. Experimental results presented in \cref{tab:supvsmim} also suggest that deep supervision offers limited value for supervised ViT on image classification.

In this paper, we revisit deep supervision for masked image modeling (MIM)~\cite{he2022masked,dong2021peco,wei2022masked,bao2021beit}, a self-supervised pre-training strategy for Vision Transformer~\cite{dosovitskiy2020image} (ViT). MIM is a simple but effective idea that trains a ViT by requiring it to predict masked inputs. Typically, MIM models adopt an encoder-decoder architecture with a masking scheme applied on image patches: the encoder (\textit{i.e.}, ViT) generates the latent representations from the last Transformer block, while the decoder predicts the reconstruction targets of the masked patches. When transferring the MIM pre-trained model to downstream tasks, the encoder is retained while the decoder is dropped. As a result, the decoder implicitly deepens the network during the pre-training stage, making the shallower layers of the encoder receives weaker informative feedback from the supervision signal.

\begin{table}[t]
\centering
\begin{tabular}{l|c|c}
\toprule
Model & Deep Supervision & IN Top-1 (\%) \\
\midrule
\multicolumn{3}{c}{\textit{Supervised learning}} \\
\midrule
Supervised Cls.  & &    81.2       \\
Supervised Cls. &   $\checkmark$  &  80.1 (\red{-1.1})    \\
Supervised Cls.$^\ddagger$ &   $\checkmark$  &  80.6 (\red{-0.6})    \\
\midrule
\multicolumn{3}{c}{\textit{Self-supervised pre-training}} \\
\midrule
MAE  &   &    82.6     \\
DeepMIM-MAE  &   $\checkmark$ &    83.4 (\textcolor{OliveGreen}{+0.8})    \\
DeepMIM$^\dagger$-MAE     &   $\checkmark$    &  83.6 (\textcolor{OliveGreen}{+1.0}) \\
\bottomrule
\end{tabular}
\caption{
\textbf{Deep supervision benefits self-supervised pretraining.}
We compare the effect of deep supervision during supervised learning and self-supervised pre-training, respectively, and report top-1 accuracy on ImageNet (IN).
The experiment adopts a DeiT-style~\cite{deit} training paradigm to optimize ViT-B on supervised image classification. We pre-train MAE and DeepMIM on MAE for 300 epochs on ViT-B.
$\ddagger$: that hyper-parameters are carefully tuned.
DeepMIM$^{\dagger}$: DeepMIM with hybrid targets (optional).
}
\label{tab:supvsmim}
\end{table}

To drive the shallower layers learn more meaningful representations, we present DeepMIM, a ViT pre-training framework with deep supervision applied to MIM pretext task: we pre-train an MAE (on ViT-B) with three extra lightweight decoders appended to the outputs of the shallow blocks (the \sixth, \eighth and \tenth blocks) of the encoder (ViT-B). Then ViT-B is finetuned on ImageNet-1K~\cite{deng2009imagenet}, following common practice~\cite{he2022masked}. 
Surprisingly, a substantial improvement (+0.8 top-1 accuracy) over the baseline (ViT-B pre-trained via the original MAE) is observed in \cref{tab:supvsmim}, highlighting the potential benefits of deep supervision in MIM pre-training.

To investigate the benefits of introducing deep supervision into MIM pre-training, we comprehensively study several factors including 1) reconstruction loss of MAE and \methodName; 2) CKA (centered kernel alignment)~\cite{kornblith2019similarity} similarities between features produced by the intermediate layers and features produced by the last layer in DeepMIM; 3) attention diversities of different attention heads. In \cref{fig:tesear_loss}, we find that DeepMIM applied to MAE attains lower
reconstruction losses across various layers in comparison with MAE. Xie \textit{et al.}~\cite{xie2022data} also reveal that the lower the reconstruction loss on the validation set is, the better performance of MIM is. In addition, we also observe that CKA scores between the last and shallow layers of \methodName always surpass those of MAE (\cref{fig:cka}) and DeepMIM learns more diverse attention heads than MAE (\cref{fig:similarity}).

Recent MIM works have studied the question: \textit{what} are appropriate reconstruction targets? Proposals have included discrete tokens~\cite{bao2021beit}, RGB pixels~\cite{he2022masked,xie2022simmim}, histograms of oriented gradients~\cite{wei2022masked} and CLIP features~\cite{peng2022beit,hou2022milan}. Our work dedicates to studying an orthogonal component of MIM pre-training: \textit{where} should the reconstruction loss be applied?  Therefore, our DeepMIM is compatible with a broad range of encode-decoder MIM models as shown in \cref{fig:tesear_acc}. Incorporating DeepMIM into various MIM models including MAE, and several other MAE variants with different reconstruction targets yield consistent improvements over the non-\methodName baselines. With parallel small decoders, \methodName only brings slightly more computation costs. The contributions of this work can be summarized as follows:
\begin{itemize}
    \item We revisit the deep supervision for MIM pre-training. In contrast to previous MIM works exploring \textit{what} form appropriate reconstruction targets should take, we focus on an orthogonal direction: \textit{where} to apply the reconstruction loss. We throughly investigate the benefits of introducing deep supervision into MIM pre-training and find that it results in lower reconstruction loss, more diverse heads, and more powerful representation capability of the shallower layers.
    \item We present an optional module termed hybrid target generator, which further boosts the performance but involves extra computational overhead. A comparison with DeepMIM-MAE and MAE is shown in \cref{tab:supvsmim}.
    
    \item DeepMIM is complementary to most existing MIM works. Extensive experiments demonstrate that MIM models equipped with \methodName significantly outperform their non-\methodName counterparts. For instance, using ViT-B, MAE with \methodName achieves 84.2 top-1 accuracy on ImageNet, outperforming MAE by +0.6 top-1 accuracy. By using \methodName with a strong reconstruction tokenizer, CLIP, we achieve state-of-the-art performance on various downstream tasks including image classification (85.6 top-1 accuracy on ImageNet-1K), object detection (52.8 AP$^{\text{box}}$ on COCO) and semantic segmentation (53.1 mIoU on ADE20K).
\end{itemize}

\section{Related Work}
\noindent\textbf{Self-supervised learning.}
Self-supervised pretraining aims to induce the model to learn transferable representations via a proxy task that does not require labels.
One approach that has seen widespread adoption is to employ a contrastive loss, modeling the similarity between different views of the same image and the dissimilarity between different images~\cite{wu2018unsupervised,chen2020simple,he2020momentum,clip,xie2021propagate,wu2018unsupervised,wang2021dense}. 
Recently, inspired by the success of BERT~\cite{devlin2018bert} in natural language processing with Masked Language Modeling, masking approaches (particularly Masked Image Modeling) have become a popular alternative for computer vision pretraining.

\noindent\textbf{Masked image modeling.}
Masked Image Modeling operates by removing a portion of the input image before it is passed to the model and training the model to reconstruct the missing content.
In the same spirit as masked language modeling, MAE~\cite{he2022masked} and SimMIM~\cite{xie2022simmim} employ raw pixels as the targets for reconstruction.
In contrast to the tokens employed as reconstruction targets in natural language processing (which exhibit rich semantics), the pixels in computer vision convey a relatively low-level signal and typically exhibit considerable redundancy.  
To inject more semantics into target tokens, several proposals for the design of an appropriate tokenizer have been put forward. 
BeiT~\cite{bao2021beit} and PeCo~\cite{dong2021peco} employ VQVAE~\cite{ramesh2021zero} to predict discrete visual vocabularies.
MaskFeat~\cite{wei2022masked} employ HOG (local gradient features) as a lightweight tokenizer with minimal parameters. 
iBOT~\cite{zhou2021ibot} and data2vec~\cite{baevski2022data2vec} use the exponential moving average of the model as their tokenizer, inspired by contrastive learning approaches. 
Our method is compatible with previous methods and brings consistent improvements when used in combination with them.

\noindent\textbf{Deep supervision.}
Deep Supervision~\cite{wang2015training,lee2015deeply,zhang2022contrastive,li2022comprehensive} was studied in the past as a tool to assist with training deep nets that could potentially mitigate gradient vanishing/exploding~\cite{hochreiter2001gradient} problems.
GoogleNet~\cite{szegedy2015going} introduces two extra supervision on intermediate layers. 
DSN~\cite{wang2015training} propose to design auxiliary supervision branches at certain intermediate layers.  
With the emergence of batch normalization~\cite{ioffe2015batch} and residual learning~\cite{he2016deep}, gradient vanishing/exploding issues appear to be less prevalent. 
It may be for this reason that deep supervision has received less interest recently.
Differently from prior work, we revisit deep supervision in the context of self-supervised learning and masked image modeling and demonstrate its value in this setting.

\section{Methodology}
\label{sec:method}
\begin{figure*}
	\centering
	\includegraphics[width=0.98\linewidth]{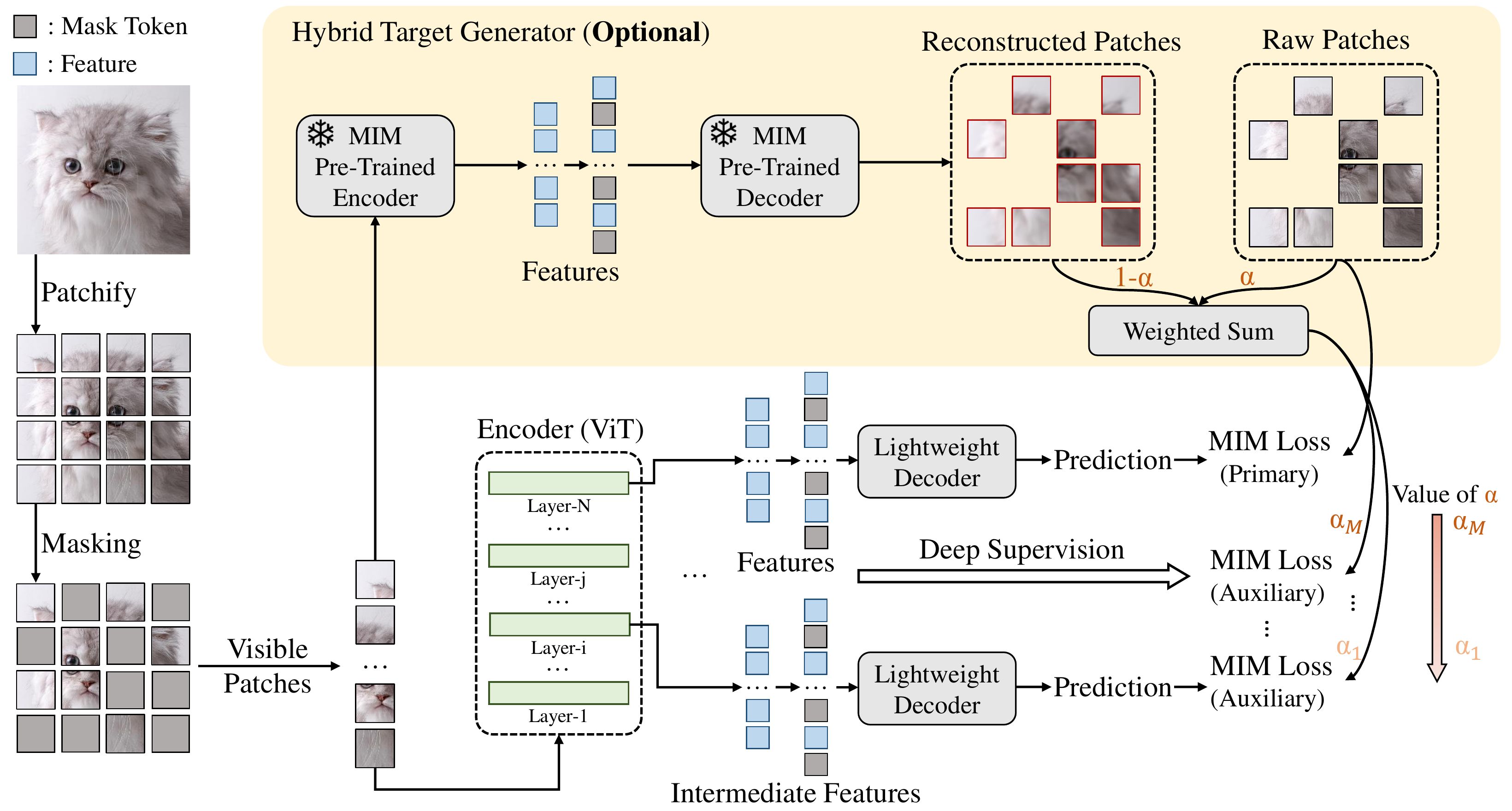}
	\caption{An overview of DeepMIM. DeepMIM applies deep supervision on intermediate features during pre-training. Each lightweight decoder is composed of 4 Transformer blocks. DeepMIM is compatible with many MIM models across a range of architectures and reconstruction targets. Using the hybrid target generator (a pre-trained MIM model) is an optional choice.}
	\label{fig:method}
\end{figure*}

As shown in~\cref{fig:method}, DeepMIM adopts an encoder-multi-decoder architecture to perform the mask-and-predict task for ViT pretraining. 
For concreteness, we use MAE~\cite{he2022masked} to illustrate our underlying approach. 
Nevertheless, DeepMIM can be applied to various masked image modeling (MIM) frameworks---we describe how at the end of this section.

\noindent\textbf{Encoder and multi-level features.} 
We adopt ViT-B with 12 Transformer blocks as the encoder $h_{\theta}$. 
Following ViT~\cite{dosovitskiy2020image}, we divide an input image $\boldsymbol{x}$ into regular non-overlapping patches. 
Then, similar to MAE~\cite{he2022masked}, we randomly mask a high proportion of patches, yielding a masked image $\widetilde{\boldsymbol{x}}$. 
We then feed $\widetilde{\boldsymbol{x}}$ into the encoder $h_{\theta}$ to produce multi-level features.

\noindent\textbf{Multi-decoder.} 
Let $g_{\xi}$ represent the default decoder attached to the last (12$^{\mathrm{th}}$) block. 
In addition to the last Transformer block, decoders are also attached to intermediate blocks. 
For ViT-B, we append three extra decoders (denoted $g_{\xi}^{1}$, $g_{\xi}^{2}$ and $g_{\xi}^{3}$) onto the \sixth, \eighth and \tenth Transformer block of the encoder $h_{\theta}$ to facilitate deep supervision. 
Each decoder is an independent 4-layer Transformer with encoded visible patches (from the last block or intermediate blocks) and masked tokens as input. Thanks to the lightweight decoders, the overall training cost of DeepMIM is slightly higher than that of MAE, i.e., \methodName and MAE take 115 and 108 training hours respectively under a 1600-epoch schedule on 32$\times$NVIDIA V100 GPUs.
We use $\boldsymbol{p}$ and $\boldsymbol{p_i}$ to denote the reconstruction prediction by $g_{\xi}$ and $g_{\xi}^{i}$, respectively.

\noindent\textbf{Progressive hybrid targets (optional).} 
The features produced by the shallower layers of the ViT are less discriminative. 
It may be beyond the capacity of these intermediate features to reconstruct the targets that are too complicated, $i.e.$, raw pixels. 
As a by-product of MIM task, a pre-trained MAE is able to recover masked images.  
Though MAE produces fuzzy reconstruction results\footnote{Image inpainting is not the objective of MAE.}, we propose to use these reconstructed images as appropriate reconstruction targets to ease the learning of the intermediate blocks. We name a pre-trained MAE as a hybrid target generator, as illustrated in Figure~\ref{fig:method}. Concretely, given a masked image $\widetilde{\boldsymbol{x}}$, we feed it into a pre-trained MAE to generate a reconstructed image, denoted $\hat{\boldsymbol{x}}$. 
The hybrid target $\boldsymbol{t}$ is generated by blending the raw image $\boldsymbol{x}$ and the reconstructed image $\hat{\boldsymbol{x}}$ with a blending ratio $\alpha$:
\begin{equation}
\label{eq:blend}
    \boldsymbol{t} = \alpha \boldsymbol{x} + (1-\alpha)\hat{\boldsymbol{x}}.
\end{equation}
Let $\boldsymbol{t}_i$ represent the reconstruction target of the decoder $g_{\xi}^{i}$. We set $\alpha$ as $0$, $1/3$ and $2/3$ for $g_{\xi}^{1}$, $g_{\xi}^{2}$ and $g_{\xi}^{3}$, respectively. 

Note that the hybrid targets are optional. Even though using hybrid targets improves fine-tuning performance, there is still additional computational overhead. We suggest to use hybrid targets only if there is an off-the-shelf hybrid target generator (a pre-trained MIM model). We use DeepMIM$^\dagger$ to denote our framework with hybrid targets. For DeepMIM (without hybrid targets), we set $\alpha=1$ for all decoders.

\noindent\textbf{Training objective.}
The overall loss is the sum of the $M+1$ $\ell$-2 reconstruction losses produced by the $M$ extra decoders and the primary decoder:
\begin{equation}
    	\mathcal{L} = \sum_{i=1}^{M} \|\mathcal{M}(\boldsymbol{t}_i)-\boldsymbol{p}_i\|_2^2 + \|\mathcal{M}(\boldsymbol{x})-\boldsymbol{p}\|_2^2, 
\label{eq:loss}
\end{equation}
where $\mathcal{M}(\cdot)$ denotes the operation that extracts the targets of the masked patches, and $M$ is the number of extra decoders. $M=3$ for ViT-B.

\noindent\textbf{\methodName is compatible with a range of MIM models.} 
\methodName can be applied to a range of MIM models with different reconstruction targets, $\textit{e.g.}$, RGB pixels~\cite{he2022masked,xie2022simmim}, discrete tokens~\cite{bao2021beit}, histograms of oriented gradients~\cite{wei2022masked}, CLIP features~\cite{hou2022milan} and DINO features~\cite{caron2021emerging}. The implementation consists of four steps:
1) utilize an off-the-shelf MIM model to produce reconstructed signals for masked images (optional);
2) generate a collection of hybrid targets by integrating raw signals and reconstructed signals with a given blending ratio (optional);
3) append extra decoders to the intermediate Transformer blocks;
4) train the MIM model equipped with \methodName through the loss defined in Eq.~\ref{eq:loss}.

\section{Discussion}
\label{sec:discuss}
We compare MAE to MAE equipped with \methodName (\methodName-MAE) with respect to the loss, layer similarity and capability of shallower blocks.
We adopt ViT-B for both methods. 
Both pre-training and fine-tuning are conducted on ImageNet-1K training set. 
Unless otherwise specified, we adopt a 300-epoch pre-training schedule.

\begin{figure}
    \includegraphics[width=1.0\linewidth]{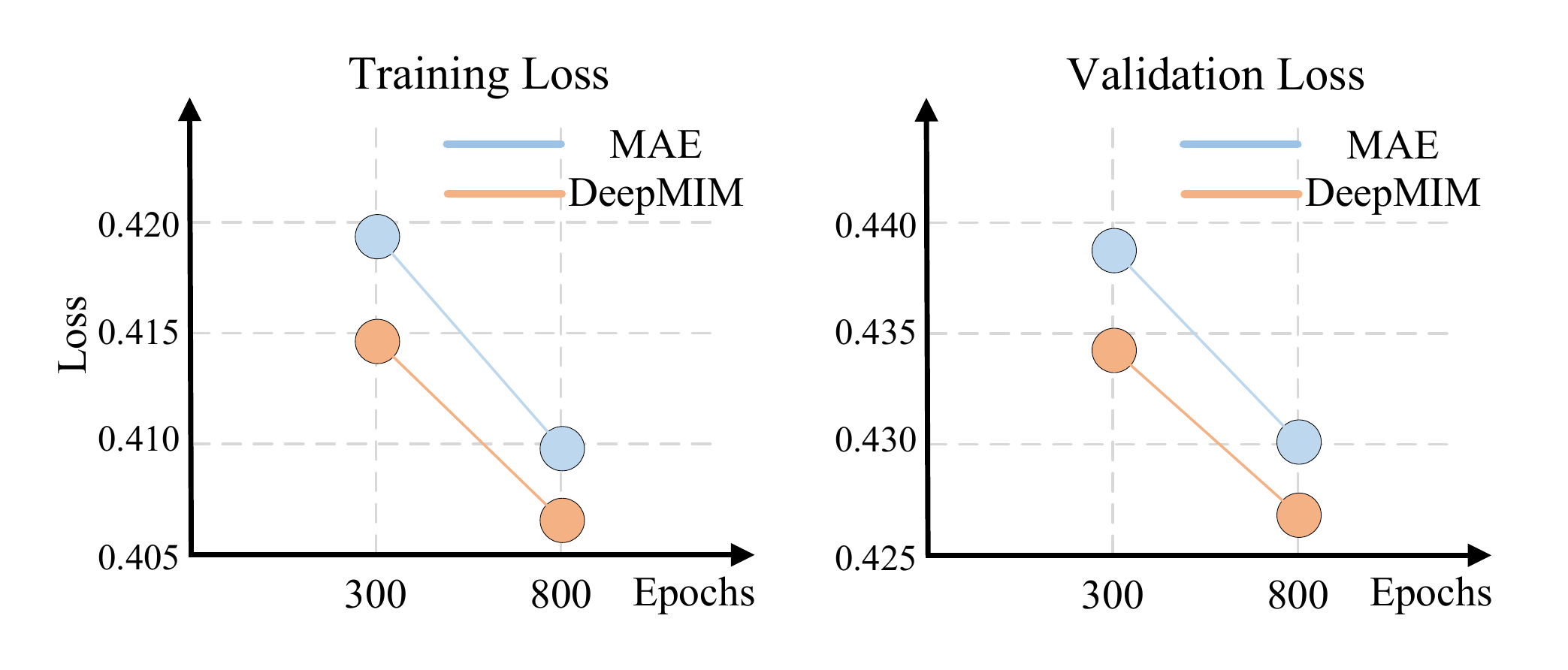}
    \vspace{-4mm}
    \caption{Comparison of the training loss (left) and the validation loss (right) of the MAE and our \methodName-MAE.
    We show only the reconstruction loss from the last block of the \methodName-MAE.}
    \vspace{-2mm}
    \label{fig:tesear_loss}
\end{figure}
\noindent\textbf{Training and validation loss.}
In \cref{fig:tesear_loss}, we plot the training loss and the validation loss of MAE and DeepMIM-MAE on ImageNet-1K, respectively. 
For a fair comparison, we only show the reconstruction loss associated with the last block of the \methodName-MAE.
We observe that \methodName-MAE lowers the training loss across different training regimes (300 and 800 epochs). 
A recent work~\cite{xie2022data} shows that validation loss in pre-training is a good indicator for how well a model will perform during fine-tuning on downstream tasks. 
For completeness, we also report validation loss, showing the identical phenomenon---\methodName-MAE achieves a lower loss than MAE.

\begin{figure}[t]
    \centering
    \begin{tabular}{c}
        \includegraphics[width=0.4\textwidth]{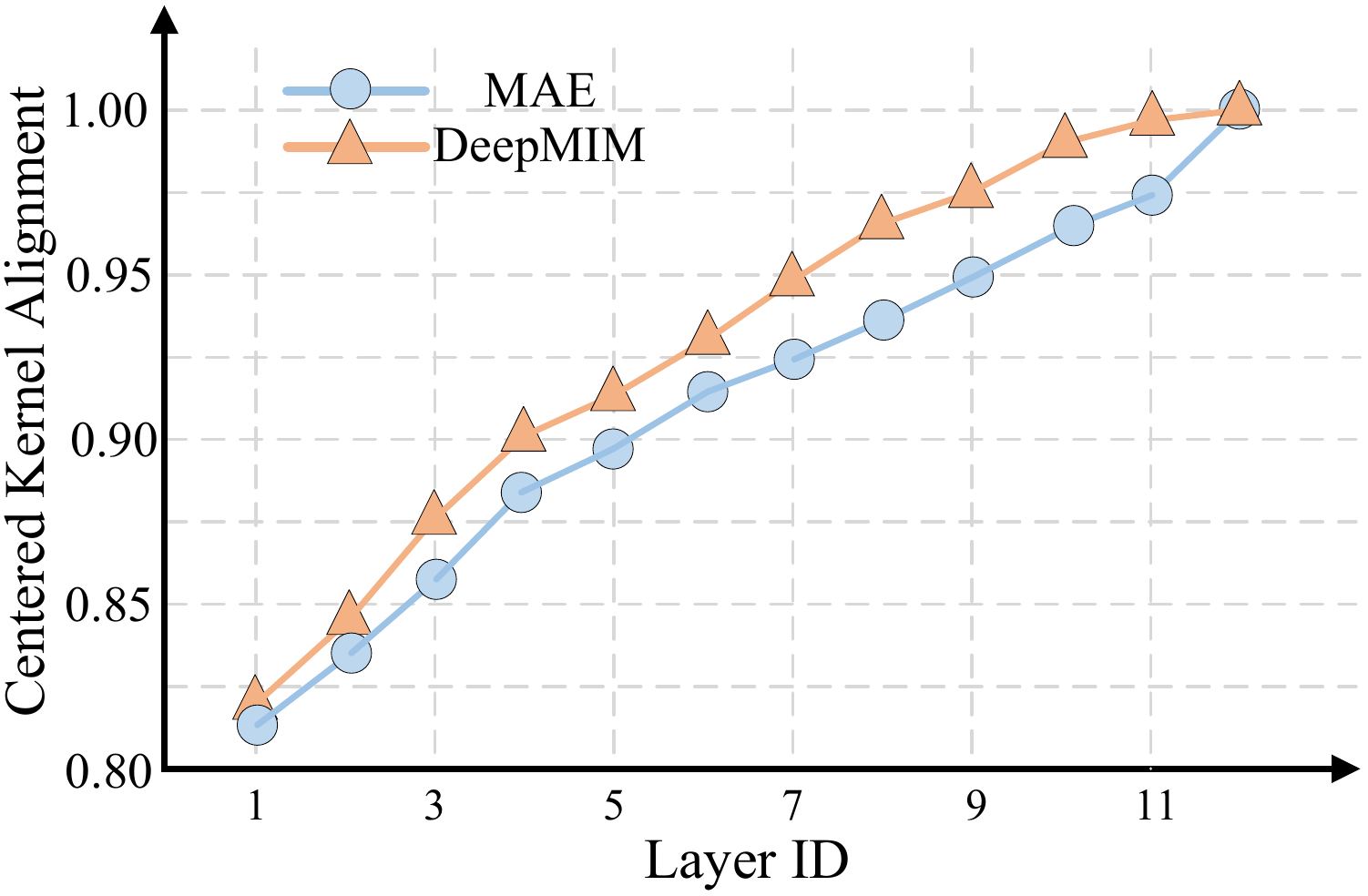}
    \end{tabular}
    \vspace{-3mm}
    \caption{We use CKA~\cite{kornblith2019similarity} to evaluate the correspondences between the feature from the last block and the features from the intermediate blocks. Blue: MAE; red: DeepMIM-MAE.}
  \label{fig:cka}
\end{figure}

\noindent\textbf{Feature similarity.} 
We employ centered kernel alignment (CKA)~\cite{kornblith2019similarity} to identify correspondences between features produced by the last block and features produced by the intermediate blocks. 
The comparison between MAE and \methodName-MAE is shown in~\cref{fig:cka}.
From the first block to the penultimate block, the CKA score of \methodName-MAE always surpasses that of MAE, suggesting that the features from the intermediate blocks of \methodName-MAE are more discriminative.

\begin{figure}[t]
    \centering
    \begin{subfigure}[b]{0.4\textwidth}
         \centering
         \includegraphics[width=\textwidth]{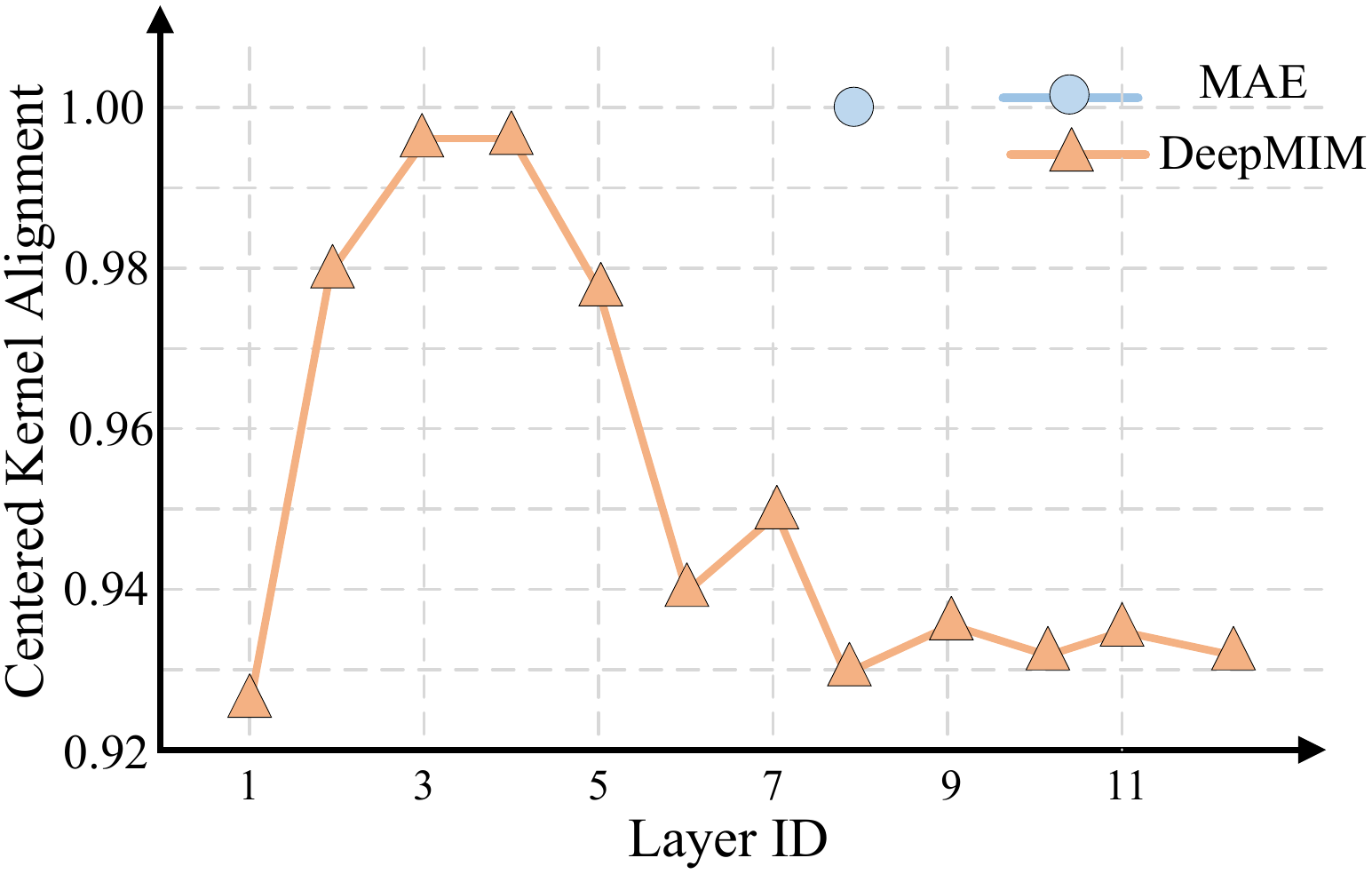}
         \caption{CKA similarity between the 8$^{\mathrm{th}}$ layer of MAE and all layers of DeepMIM.}
         \label{fig:cross_cka_a}
     \end{subfigure}
     \hfill
     \begin{subfigure}[b]{0.4\textwidth}
         \centering
         \includegraphics[width=\textwidth]{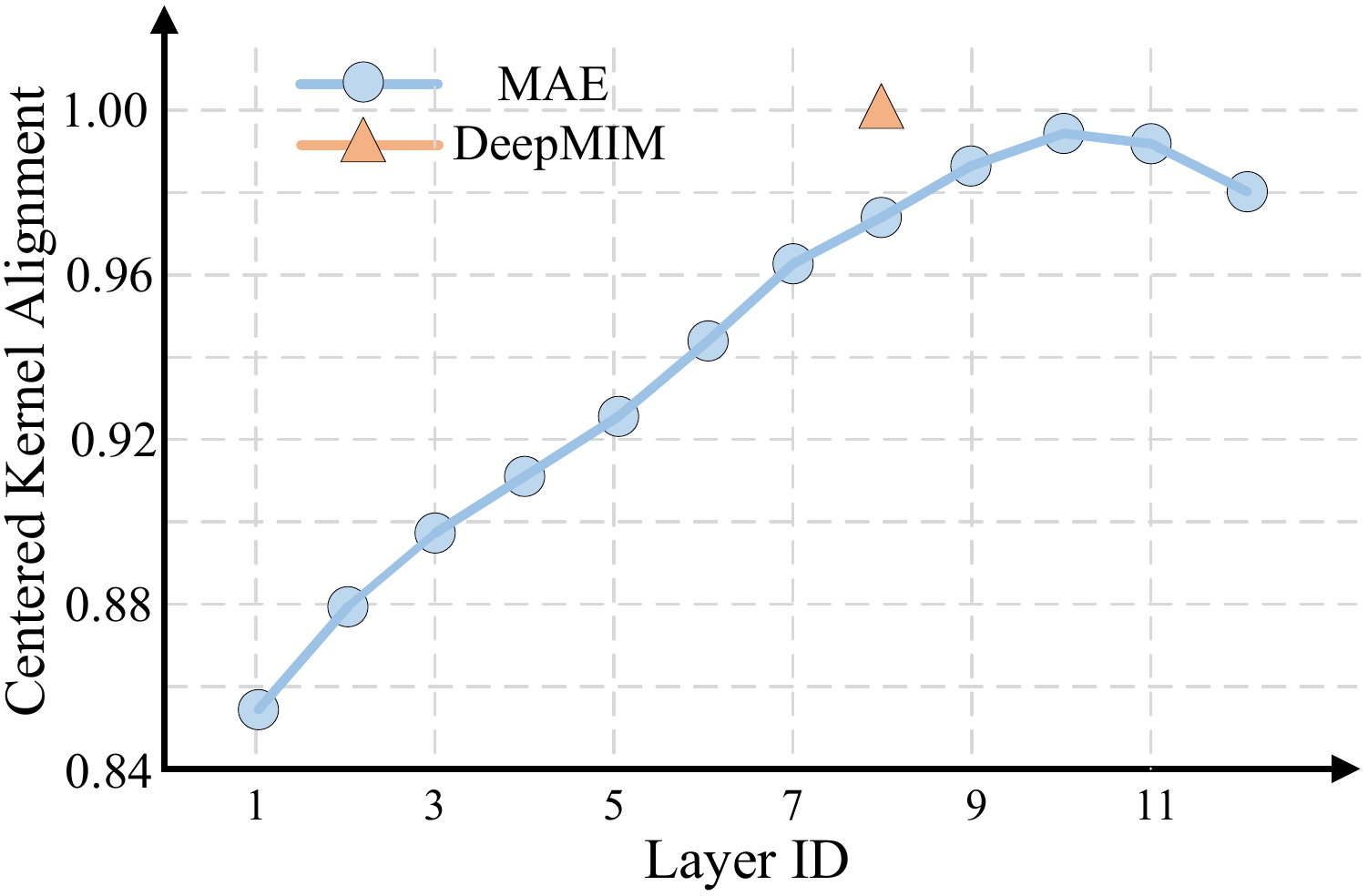}
         \caption{CKA similarity between the 8$^{\mathrm{th}}$ layer of DeepMIM and all layers of MAE.}
          \label{fig:cross_cka_b}
     \end{subfigure}
     \vspace{-3mm}
    \caption{Cross feature similarities evaluated by CKA.
    }
  \label{fig:cross_cka}
\end{figure}

\noindent\textbf{Cross feature similarity.}
We use CKA to calculate similarities between features from the \eighth block of MAE and features from all blocks of \methodName-MAE as shown in~\cref{fig:cross_cka_a}. 
We also plot a reversed version in~\cref{fig:cross_cka_b}.
The MAE's intermediate (\eighth block) features exhibit greatest alignment with \methodName-MAE's shallower block features (3$^{\mathrm{th}}$ and 4$^{\mathrm{th}}$).
In contrast, the intermediate (\eighth block) features of \methodName-MAE align more closely with features from the deeper blocks (9$^{\mathrm{th}}$ and \tenth) of MAE. 
This study also indicates that DeepMIM significantly strengths the discriminative power of the features from the shallower blocks. 

\noindent\textbf{Attention Head Diversity.}
\begin{figure}
\begin{tabular}{cc}
\hspace{-2mm}
     \includegraphics[width=0.48\linewidth]{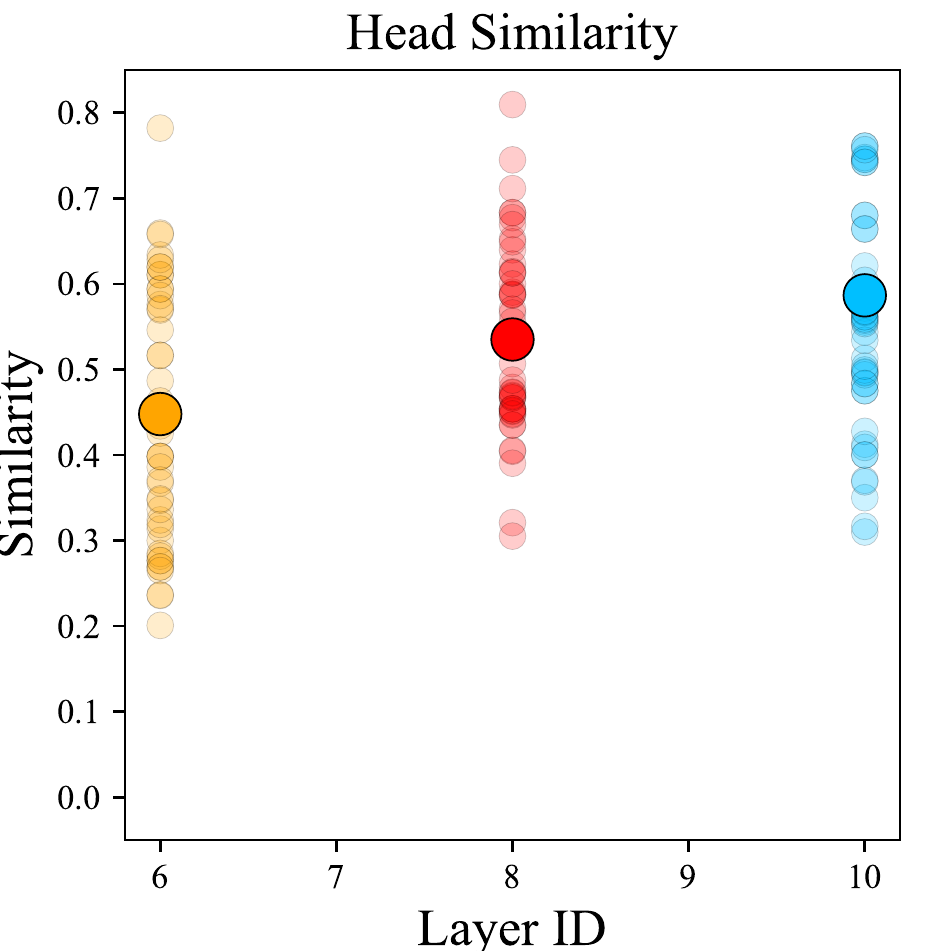}& 
     \hspace{-3mm}\includegraphics[width=0.48\linewidth]{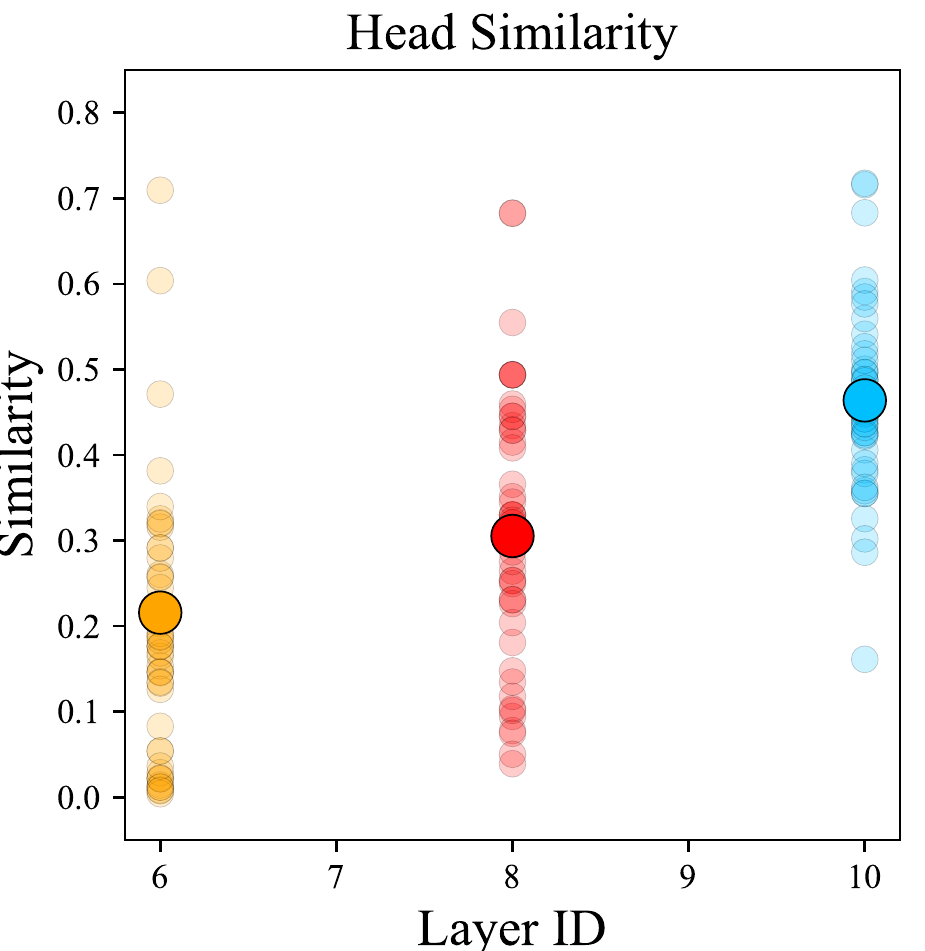} \\
\end{tabular}
    \caption{Comparison of the head cosine similarity of the MAE (left) and our \methodName-MAE (right) across different layers. Large dots denote the averaged similarities.}
    \label{fig:similarity}
\end{figure}
We calculate the cosine similarities between different attention heads to explore the head diversities. According to~\cite{xie2022revealing}, more diverse heads indicate a stronger capacity for representation. As shown in \cref{fig:similarity}, we plot the cosine similarities of different attention heads across various layers of MAE and \methodName. Our DeepMIM yields more diverse heads in comparison with MAE.

\begin{figure}[t]
    \centering
    \begin{tabular}{c}
        \includegraphics[width=0.4\textwidth]{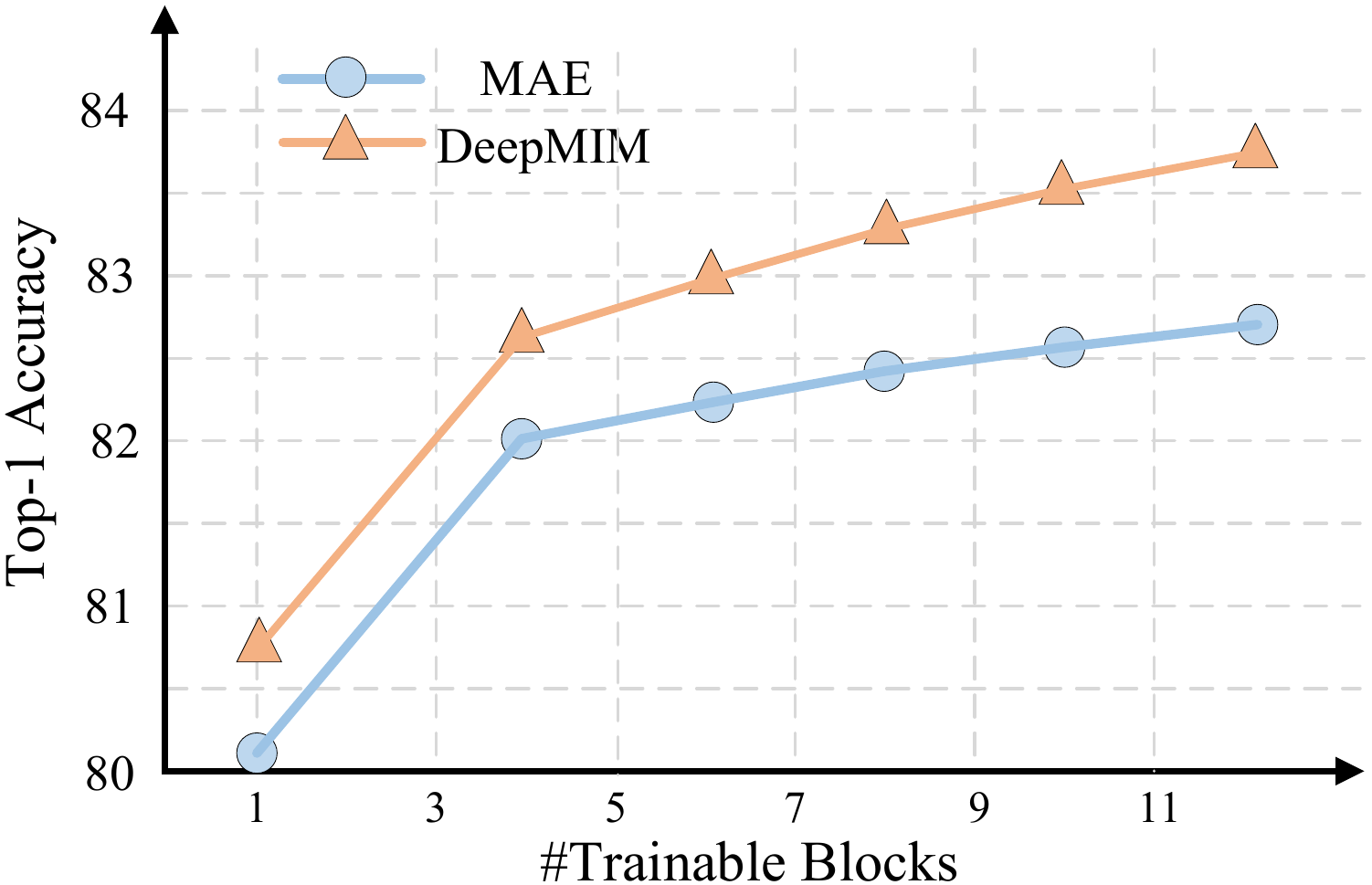}
    \end{tabular}
    \vspace{-3mm}
    \caption{When varying the number of trainable blocks from 1 to 12, DeepMIM-MAE consistently outperforms MAE by apparent margins.}
  \label{fig:partial}
\end{figure}

\noindent\textbf{Fine-tuning the last-K blocks.}
To further assess the quality of the features from the shallower blocks, we freeze a subset of shallow blocks and fine-tune the remaining ones. 
For this experiment, we report top-1 accuracy on ImageNet.
As shown in~\cref{fig:partial}, when the number of trainable blocks is varied from 1 (only the last block is trainable) to 12 (all blocks are trainable), \methodName-MAE consistently outperforms MAE by a significant margin.

\begin{figure}[t]
    \centering
    \begin{tabular}{c}
        \includegraphics[width=0.4\textwidth]{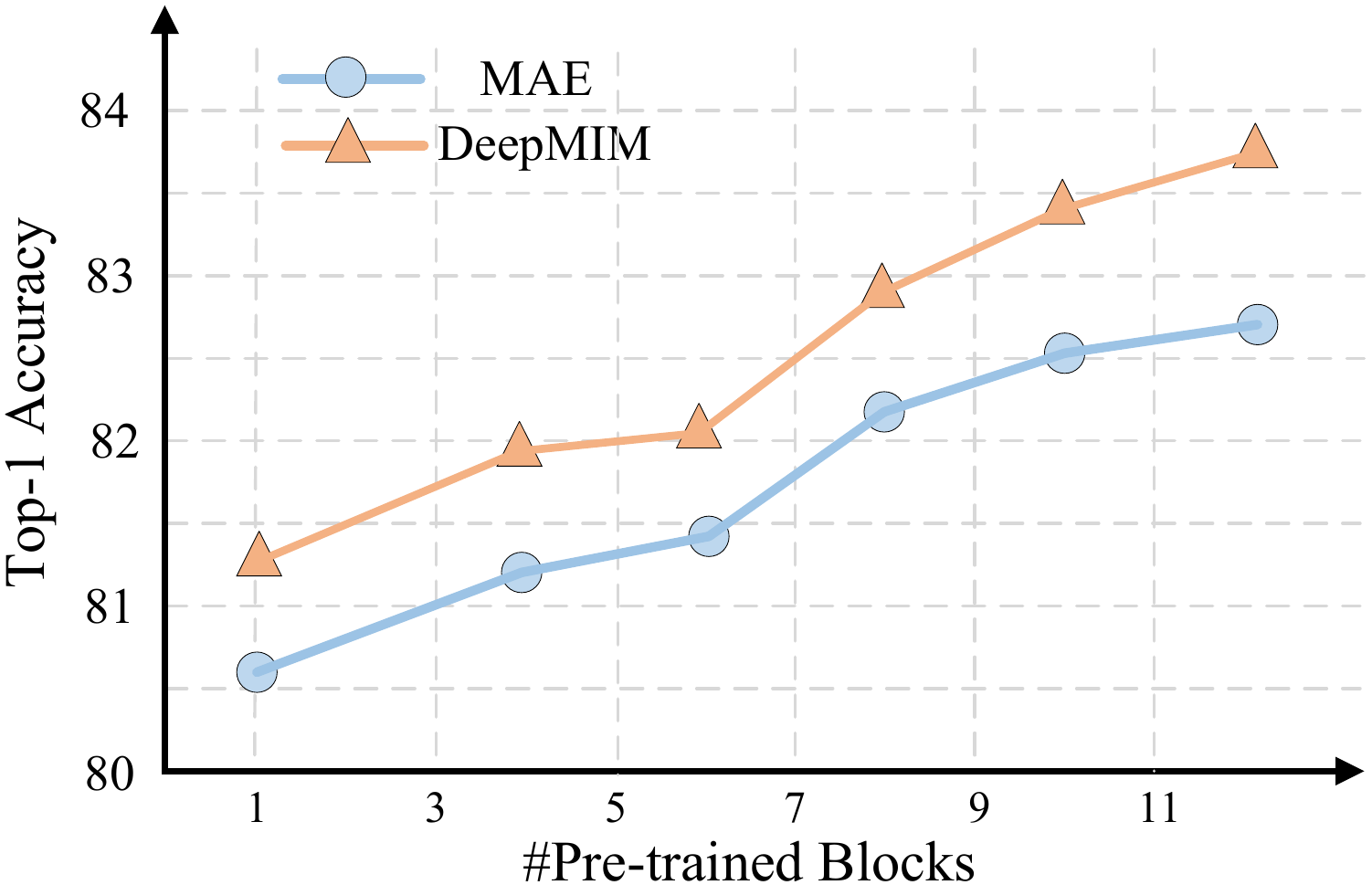}
    \end{tabular}
    \vspace{-3mm}
    \caption{End-to-End finetune on ImageNet-1K. 
    Start k layers are pretrained by MAE/Ours, and rest layers are randomly initialized. }
    \vspace{-3mm}
  \label{fig:rand}
\end{figure}

\noindent\textbf{Randomly initialising the last-K blocks when fine-tuning.}
For this experiment, we first randomly initialize the last-K blocks of a pre-trained ViT-B, then the ViT-B is fine-tuned on ImageNet in an end-to-end manner.
\cref{fig:rand} shows a comparison between MAE and \methodName-MAE.
We observe that \methodName-MAE consistently outperforms MAE in each case, demonstrating that good representations at shallow blocks are advantageous for the learning of deeper blocks particularly when they are randomly initialized.

\section{Experiments}
\noindent\textbf{Implementation Details.}
We adopt ViT-B/16 as our backbone with input size of $224\times 224$. 
The ImageNet-1K image classification dataset is used for both pre-training and fine-tuning~\cite{deng2009imagenet}. 
We apply our \methodName on MAE with RGB pixels~\cite{he2022masked} and CLIP features~\cite{radford2021learning}. 
For comparison with state-of-the-art methods, we pre-train \methodName for 300 and 1600 epochs.
We follow the same training recipe of MAE~\cite{he2022masked}, more details can be found in supplementary materials. We adopt the hybrid target generator (a pre-trained MIM model) if the reconstruction targets are pixels and detach it when using other reconstruction targets, since there are lots of off-the-shelf pre-trained MIM models with pixels as the reconstruction targets (e.g. from Huggingface). We use DeepMIM$^\dagger$ to indicate \methodName with hybrid targets.

\subsection{Main Results}

\begin{table}[t]
	\centering
	\small
	\begin{tabular}{lccccc}
		\toprule
		Method & Epochs    & Target & Top-1 \\
		\midrule
		\multicolumn{4}{c}{\textit{ViT-B}} \\
		BeiT~\cite{bao2021beit}& 300 & DALL-E& 83.2  \\
		CIM~\cite{fang2022corrupted}& 300 & Pixel& 83.3 \\
		CAE~\cite{chen2022context}& 300  & Momentum & 83.6 \\
		MaskFeat~\cite{wei2022masked} & 300 & HOG & 83.6 \\
		iBoT~\cite{zhou2021ibot}       &  400 & Momentum  & 83.8  \\
		PeCo~\cite{dong2021peco} & 300  & Codebook & 84.1  \\
		\midrule
		MAE~\cite{he2022masked}   & 300 & Pixel & 82.6  \\
  \methodName-MAE & 300   & Pixel & 83.4 \\
		DeepMIM$^\dagger$-MAE & 300   & Pixel & 83.6 \\
		\midrule
		MAE-CLIP$^\star$~\cite{he2022masked}   & 300 & CLIP& 83.8  \\
		\methodName-MAE-CLIP & 300  & CLIP & 84.8 \\
		\midrule
		\midrule
		\multicolumn{4}{c}{\textit{ViT-B + longer pre-training}} \\
		DINO~\cite{caron2021emerging}    & 1600  & Momentum & 82.8 \\
		BEiT~\cite{bao2021beit}          & 800  & DALL-E & 83.2 \\
		SimMIM~\cite{xie2022simmim}      & 800  & Pixel   & 83.8 \\
		SIM~\cite{tao2022siamese}        & 1600  & Momentum & 83.8 \\
		CAE~\cite{chen2022context}& 1600  & Momentum & 83.9 \\
		MaskFeat~\cite{wei2022masked}    & 1600  & HOG & 84.0  \\
		iBOT~\cite{zhou2021ibot}         & 1600  & Momentum & 84.0 \\
		PeCo~\cite{dong2021peco}         & 800  & Codebook & 84.5 \\
		\midrule
		MAE~\cite{he2022masked}         & 1600 & Pixel & 83.6 \\
  		DeepMIM-MAE & 1600   & Pixel & 84.0 \\
		DeepMIM$^\dagger$-MAE & 1600   & Pixel & 84.2 \\
		\midrule
		MAE-CLIP$^\star$ & 1600  & CLIP & 84.8 \\
		DeepMIM-MAE-CLIP & 1600  & CLIP & 85.6 \\
		\bottomrule
	\end{tabular}
 \vspace{-3mm}
	\caption{
 Comparison with previous self-supervised pre-training methods on ViT-B/16. 
 We report top-1 accuracy on ImageNet-1K. 
 $\star$: reproduced result. MAE-CLIP: an MAE variant with CLIP features as targets.}
  \vspace{-1mm}
	\label{tab:ImageNet-1k}
\end{table}

\noindent\textbf{Image Classification.} 
In~\cref{tab:ImageNet-1k}, we compare the fine-tuning results of various self-supervised pre-training methods on ImageNet-1K. 
We report the top-1 accuracy on the ImageNet validation set. 
Under a 300-epoch pre-training schedule, DeepMIM$^\dagger$-MAE (83.6) outperforms the MAE baseline (82.6) by +1.0. 
With a more powerful tokenizer like CLIP, \methodName-MAE-CLIP achieves 84.8 top-1 accuracy, outperforming the baseline of MAE-CLIP by +1.0. 

When conducting a longer pre-training schedule (1600 epochs), DeepMIM$^\dagger$-MAE (84.2) surpasses MAE (83.6) by +0.6. \methodName-MAE-CLIP (85.6) outperforms MAE-CLIP (84.8) by +0.8, achieving a new state-of-the-art.

\noindent\textbf{Object Detection.}
Following common practice, we use the COCO~\cite{lin2014microsoft} benchmark to evaluate the transferability of \methodName to the object detection task.
We employ \methodName pre-trained ViT-B/16 as the backbone and Mask R-CNN~\cite{he2017mask} as the detector.
We report AP$^{\text{box}}$ AP$^{\text{mask}}$ in~\cref{tab:coco_det}. 
DeepMIM$^\dagger$-MAE outperforms the MAE baseline by +1.3 AP$^{\text{box}}$ and +0.3 AP$^{\text{mask}}$. \methodName with MAE-CLIP achieves state-of-the-art performance of 52.8 AP$^{\text{box}}$ and 46.0 AP$^{\text{mask}}$.
\begin{table}
		\centering
		\setlength{\tabcolsep}{8pt}
		\begin{tabular}{l|c|c|c}
			\toprule
			Method & Epochs  & AP$^{\text{box}}$ & AP$^{\text{mask}}$ \\
			\midrule
			PeCo~\cite{dong2021peco} & 800 & 44.9 & 40.4 \\
			MoCo-v3~\cite{chen2021mocov3} & 300  & 47.9 & 42.7 \\
                BEiT~\cite{bao2021beit}& 800 &  49.8 & 44.4 \\
			SIM~\cite{tao2022siamese} & 1600 & 49.1 & 43.8 \\
			iBOT~\cite{zhou2021ibot} & 1600 & 51.2 & 44.2\\
			CAE~\cite{chen2022context} & 1600 & 50.1 & 44.0 \\
			MAE~\cite{he2022masked}  & 1600 & 50.3 & 44.9 \\
   \midrule
			DeepMIM$^\dagger$-MAE & 1600 & 51.6 & 45.2  \\
			\methodName-MAE-CLIP & 1600 & 52.8 & 46.0  \\
			\bottomrule
			
		\end{tabular}
   \vspace{-3mm}
		\caption{COCO~\cite{lin2014microsoft} object detection and instance segmentation using Mask R-CNN~\cite{he2017mask}. 
		}
		\label{tab:coco_det}
\end{table}

\begin{table}[t]
	\centering
	\renewcommand\arraystretch{1.1}
		\begin{tabular}{l|c|c}
			\toprule
			Method & Epochs & mIoU\\
			\midrule
			CIM~\cite{fang2022corrupted} & 300 &  43.5 \\
			BEiT~\cite{bao2021beit} & 800  & 47.1\\
			MoCo-v3~\cite{chen2021mocov3} & 300 & 47.3 \\
			DINO~\cite{caron2021emerging} & 400& 47.2 \\
			PeCo~\cite{dong2021peco} & 800 & 48.5\\
			iBOT~\cite{zhou2021ibot} & 1600 & 50.0\\
   CAE~\cite{chen2022context} & 1600&  50.2 \\
			MAE~\cite{he2022masked} & 1600 &48.1 \\
                \midrule
			DeepMIM$^\dagger$-MAE & 1600 &49.5 \\
			\methodName-MAE-CLIP & 1600 & 53.1 \\
			\bottomrule
		\end{tabular}
  \vspace{-3mm}
		\caption{ADE20K~\cite{ade20k} semantic segmentation using UperNet~\cite{xiao2018unified}}
  \vspace{-3mm}
		\label{tab:seg}
	\end{table}
	
\noindent\textbf{Semantic Segmentation.}
We also transfer the \methodName pre-trained ViT-B/16 to semantic segmentation on the ADE20K~\cite{ade20k} benchmark.
We use UperNet~\cite{xiao2018unified} for a fair comparison with previous methods.
We report mean intersection over union (mIoU) for each model in~\cref{tab:seg}.
DeepMIM$^\dagger$-MAE outperforms its counterpart by +1.4, and \methodName-MAE-CLIP achieves the state-of-the-art performance of 53.1 mIoU.

\noindent\textbf{Video Classification.}
This transfer experiment is conducted on Kinetics-400~\cite{kay2017kinetics} benchmark using ViT-B. 
We apply DeepMIM to VideoMAE~\cite{tong2022videomae}, and report top-1 accuracy on Kinetics-400 validation set. 
We follow the same pre-training and fine-tuning settings of VideoMAE and pre-train \methodName-VideoMAE for 800 epochs for a fair comparison with VideoMAE. 
The results are reported in~\cref{tab:k400}.
\methodName-VideoMAE improves over the baseline VideoMAE by +1.2.

\noindent\textbf{Robustness Evaluation on Out-of-domain Datasets.}
We evaluate \methodName-MAE on three out-of-domain datasets~\cite{imageneta,imagenetr,imagenetc}: ImageNet-A (natural adversarial examples), ImageNet-R (semantic shifts), and ImageNet-C (image corruptions). We report top-1 accuracy on ImageNet-A/R and mCE error on ImageNet-C. The results are shown in \cref{tab:robust}, our \methodName significantly outperforms MAE baseline by large margins.

\begin{table}[t!]
	\centering
	\begin{tabular}{l|c|c}
		\toprule
		Method &Pre-Data  & Top-1 Acc. \\
		\midrule
		
		NL I3D~\cite{nonlocal} &  \multirow{4}{*}{ImageNet-1K}   & 77.3   \\
		
		TANet~\cite{tanet} &   & 79.3 \\
		TDN$_{En}$~\cite{tdn} &  & 79.4    \\
		Video Swin~\cite{liu2021video}&  &  80.6   \\
		\midrule
		TimeSformer~\cite{timesformer} & \multirow{2}{*}{ImageNet-21K} & 78.3 \\
		Motionformer~\cite{motionformer} &     & 80.2 \\
		\midrule
		VideoMAE~\cite{tong2022videomae} &  \multirow{2}{*}{Kinetics-400}  & 80.0  \\
		\methodName-VideoMAE  &&  81.2 \\
		\bottomrule
	\end{tabular}
 \vspace{-3mm}
	\caption{Kinetics-400~\cite{kay2017kinetics} video classification. }
  \vspace{-2mm}
	\label{tab:k400}
\end{table}

\begin{table}[t]
\centering
\begin{tabular}{l|c|c|c}
\toprule
Method& IN-A$\uparrow$ &IN-R$\uparrow$&IN-C $\downarrow$ \\
\midrule
DeiT~\cite{deit} &  25.8      & 45.4 & 36.8\\
MAE~\cite{he2022masked}  &   33.6  &50.0& 37.8     \\
\methodName-MAE-CLIP &   52.0  &64.5 &  33.0    \\
\bottomrule
\end{tabular}
\vspace{-3mm}
\caption{Robustness evaluation on out-of-domain datasets.}
\vspace{-2mm}
\label{tab:robust}
\end{table}

\subsection{Analysis and Ablation Study}
For all ablation studies, we adopt \methodName-MAE with a 300-epoch pre-training schedule and a 100-epoch fine-tuning schedule on ImageNet-1K (reporting top-1 accuracy). 
We employ ViT-B as the backbone.

\begin{table}[t]
	\centering
	\begin{tabular}{c|c|c}
		\toprule
		Deep Supervision    & Hybrid Target   & Top-1 Acc. \\
		\midrule
		   &          &   82.6  \\
		 \checkmark &          &   83.4  \\
		 \checkmark  &     \checkmark      &   83.6  \\
		\bottomrule
	\end{tabular}
 \vspace{-3mm}
	\caption{Ablation study on the effectiveness of deep supervision and hybrid targets.}
 \vspace{-3mm}
	\label{tab:component}
\end{table}

\noindent\textbf{The effectiveness of deep supervision and progressive hybrid targets.}
\methodName presents two techniques:
1) appending extra decoders to the intermediate blocks of the encoder to enable deep supervision for MIM pre-training;
2) utilizing progressive hybrid targets as the reconstruction targets for intermediate features. 
\cref{tab:component} demonstrates the effectiveness of each proposed technique.

\noindent\textbf{Deep Supervision.} 
\methodName appends extra decoders onto the intermediate blocks of the encoder.
Here we explore: 1) where to apply deep supervision; 2) how many blocks should be involved.
We do not use progressive hybrid targets in this study for efficiency.
The results are shown in~\cref{tab:stages}.
We compare several variants with MAE baseline, which uses a single decoder on the output of the last Transformer block.
We conclude that: 1) Regardless of how many blocks are used for deep supervision, all variations outperform the baseline.
2) Deep supervision should be applied to the appropriate blocks.
For instance, the variant with \{3$^{\mathrm{rd}}$, 6$^{\mathrm{th}}$, 9$^{\mathrm{th}}$, 12$^{\mathrm{th}}$\} blocks involved deep supervision performs worse than that involving \{6$^{\mathrm{th}}$, 9$^{\mathrm{th}}$, 12$^{\mathrm{th}}$\} blocks.
3) The configuration which involves \{6$^{\mathrm{th}}$, 8$^{\mathrm{th}}$, 10$^{\mathrm{th}}$, 12$^{\mathrm{th}}$\} blocks yields the best result, outperforming the MAE baseline by +0.8.

\begin{table}[]
	\centering
	\begin{tabular}{c|c|c}
		\toprule
		Intermediate Block    & Last Block (12$^{\mathrm{th}}$)     & Top-1 Acc. \\
		\midrule
		    -     &  \checkmark         &   82.6  \\
		\midrule
		  6$^{\mathrm{th}}$ & \checkmark      &  82.8  \\
		8$^{\mathrm{th}}$  &\checkmark      & 82.9\\
		9$^{\mathrm{th}}$ & \checkmark      &  82.9   \\
		\midrule
		6$^{\mathrm{th}}$, 9$^{\mathrm{th}}$ & \checkmark  &  83.1   \\
		  8$^{\mathrm{th}}$, 10$^{\mathrm{th}}$      & \checkmark  &  83.1 \\
		\midrule
		3$^{\mathrm{th}}$, 6$^{\mathrm{th}}$, 9$^{\mathrm{th}}$& \checkmark &  83.0   \\
		4$^{\mathrm{th}}$, 6$^{\mathrm{th}}$, 8$^{\mathrm{th}}$ & \checkmark &   83.2  \\
		6$^{\mathrm{th}}$, 8$^{\mathrm{th}}$, 10$^{\mathrm{th}}$ & \checkmark&  83.4  \\
		\bottomrule
	\end{tabular}
  \vspace{-3mm}
	\caption{Study of deep supervision using \methodName-MAE. First row is an MAE baseline.}
 \vspace{-2mm}
	\label{tab:stages}
\end{table}

\noindent\textbf{Progressive hybrid targets.} 
In~\cref{sec:method}, we propose to generate hybrid targets for different intermediate blocks by blending raw signals and the reconstructed signals with a blending ratio of $\alpha$ that controls the ratio between them (see~\cref{eq:blend}). 
Here we study using different blending ratios for the configuration that involves deep supervision on the \{4$^{\mathrm{th}}$, 8$^{\mathrm{th}}$, 10$^{\mathrm{th}}$, 12$^{\mathrm{th}}$\} blocks in~\cref{tab:genraw}.
We observe a slight improvement for the configuration where the targets of the \{4$^{\mathrm{th}}$, 8$^{\mathrm{th}}$, 10$^{\mathrm{th}}$\} blocks are pure reconstructed signals. 
Adopting progressive hybrid targets ($\alpha=0,1/3,2/3,1$ for the 6$^{\mathrm{th}}$, 8$^{\mathrm{th}}$, 10$^{\mathrm{th}}$ and 12$^{\mathrm{th}}$ blocks) yields the best performance, achieving 83.6\% top-1 accuracy.

\begin{table}[t]
	\centering
	\begin{tabular}{c|cccc|c}
		\toprule
		\multirow{2}{*}{Config ID}  & \multicolumn{4}{c|}{Blending Ratio ($\alpha$)} & \multirow{2}{*}{Top-1 Acc.} \\
		 &   6$^{\mathrm{th}}$    &8$^{\mathrm{th}}$ &10$^{\mathrm{th}}$     & 12$^{\mathrm{th}}$      &    \\
		\midrule
		1 & 1 & 1 & 1 & 1 & 83.5 \\
		2 & 0 & 0 & 0 & 1 & 83.4 \\
		3 & 1/2 & 1/2 & 1/2 & 1/2 & 83.4\\
		4 & 0 & 1/3 & 2/3 & 1& 83.6 \\
		\bottomrule
	\end{tabular}
 \vspace{-3mm}
	\caption{Study of progressive hybrid targets. We vary the blending ratio $\alpha$ of the \{6$^{\mathrm{th}}$, 8$^{\mathrm{th}}$, 10$^{\mathrm{th}}$, 12$^{\mathrm{th}}$\} blocks.}
  \vspace{-2mm}
	\label{tab:genraw}
\end{table}

\noindent\textbf{Shared decoder or independent decoder.}
By default, \methodName appends independent decoders at intermediate blocks.
We now examine the possibility of employing a single shared decoder for different blocks as shown in~\cref{tab:shared}.
Using a shared decoder significantly degrades the performance, possibly because the feature distributions of different blocks vary significantly from one to another.

\begin{table}[]
	\centering
	\begin{tabular}{c|c|c}
		\toprule
		 Block  & Decoder   & Top-1 Acc. \\
		\midrule
		 6$^{\mathrm{th}}$, 12$^{\mathrm{th}}$ & Shared     &  82.1  \\
		 6$^{\mathrm{th}}$, 12$^{\mathrm{th}}$ & Independent &    82.9  \\
		\midrule
		6$^{\mathrm{th}}$, 8$^{\mathrm{th}}$, 10$^{\mathrm{th}}$, 12$^{\mathrm{th}}$ &Shared   &  81.4   \\
		6$^{\mathrm{th}}$, 8$^{\mathrm{th}}$, 10$^{\mathrm{th}}$, 12$^{\mathrm{th}}$ & Independent &    83.4   \\
		\bottomrule
	\end{tabular}
 \vspace{-3mm}
	\caption{Comparison of shared or Independent decoder in deep supervision. All supervision is the raw patches except without hybrid targets.}
  \vspace{-2mm}
	\label{tab:shared}
\end{table}

\noindent\textbf{\methodName with different reconstruction targets.}
Differently from reconstruction targets in natural language processing with rich semantics, reconstruction targets in computer vision are low-level pixels~\cite{he2022masked,xie2022simmim}. 
Recent works~\cite{ren2023tinymim,he2022masked,dong2021peco} have explored more semantic reconstruction targets, \textit{e.g.}, CLIP features~\cite{radford2021learning}.
We apply \methodName to different MAE variants whose targets are RGB pixels~\cite{he2022masked}, HOG~\cite{wei2022masked}, DINO features~\cite{caron2021emerging}, discrete tokens generated by perceptual codebook~\cite{dong2021peco} and CLIP features~\cite{radford2021learning}. 
As shown in~\cref{tab:target}, without bells and whistles, \methodName consistently outperforms its counterpart in each comparison.

\noindent\textbf{Training Complexity.} Compared with MAE (on ViT-B), \methodName adopts four lightweight decoders and DeepMIM$^\dagger$ extra utilizes a hybrid target generator (a pre-trained MIM model). In contrast to the original MAE decoder, which uses eight Transformer blocks, each lightweight decoder in our DeepMIM is made up of four blocks. Besides, we suggest to use the hybrid target generator only if a pre-trained MIM model is available. Therefore, The extra cost of DeepMIM$^\dagger$ over DeepMIM is the inference of generating hybrid targets. The pre-training cost of \methodName and \methodName$^\dagger$ are 115 hours and 119 hours respectively, which are slightly higher than 108 hours of MAE under a 1600-epoch pre-training schedule on 32$\times$NVIDIA V100 GPUs.

\begin{table}[]
	\centering
	\begin{tabular}{l|c|c}
		\toprule
		Method    & Target & Top-1 Acc. \\
		\midrule
		MAE  & Pixel & 82.6 \\
		DeepMIM  & Pixel & 83.4 (\textcolor{OliveGreen}{+0.8}) \\
		\midrule
		MAE  & HOG~\cite{wei2022masked} & 83.4 \\
		DeepMIM  & HOG~\cite{wei2022masked} & 83.9 (\textcolor{OliveGreen}{+0.5}) \\
		\midrule
		MAE  & DINO Features~\cite{caron2021emerging} & 83.8 \\
		DeepMIM  & DINO Features~\cite{caron2021emerging} & 84.4 (\textcolor{OliveGreen}{+0.6}) \\
		\midrule
		MAE & Perceptual Codebook~\cite{dong2021peco} & 83.7 \\
		DeepMIM & Perceptual
		Codebook~\cite{dong2021peco} & 84.3 (\textcolor{OliveGreen}{+0.6}) \\
		\midrule
		MAE & CLIP Features~\cite{radford2021learning} & 83.8 \\
		DeepMIM & CLIP Features~\cite{radford2021learning} & 84.8 (\textcolor{OliveGreen}{+1.0}) \\
		\bottomrule
	\end{tabular}
   \vspace{-3mm}
	\caption{Comparison of our method with MAE models with different reconstruction targets. All models are pre-trained for 300 epochs. Our methods bring consistent improvements. }
  \vspace{-2mm}
	\label{tab:target}
\end{table}

\section{Conclusion}
In this paper, we shift our focus from designing reconstruction targets to the question of where to apply the reconstruction loss. 
We find intermediate features from shallower Transformer blocks also have predictive power for reconstruction and that improving these features during training improves the quality of learned representation over the whole model. 
Therefore, we propose \methodName that applies deep supervision on the intermediate features with extra decoders and hybrid targets to provide appropriate supervision for less discriminative intermediate features. 
Our experiments demonstrate that \methodName is compatible with a range masked image modeling frameworks and produces consistent improvements over strong baselines.

{\small
\bibliographystyle{ieee_fullname}
\bibliography{egbib}

\begin{thebibliography}{10}\itemsep=-1pt

\bibitem{baevski2022data2vec}
Alexei Baevski, Wei-Ning Hsu, Qiantong Xu, Arun Babu, Jiatao Gu, and Michael
  Auli.
\newblock Data2vec: A general framework for self-supervised learning in speech,
  vision and language.
\newblock {\em arXiv preprint arXiv:2202.03555}, 2022.

\bibitem{bao2021beit}
Hangbo Bao, Li Dong, and Furu Wei.
\newblock Beit: Bert pre-training of image transformers.
\newblock {\em arXiv preprint arXiv:2106.08254}, 2021.

\bibitem{timesformer}
Gedas Bertasius, Heng Wang, and Lorenzo Torresani.
\newblock Is space-time attention all you need for video understanding?
\newblock In {\em ICML}, 2021.

\bibitem{caron2021emerging}
Mathilde Caron, Hugo Touvron, Ishan Misra, Herv{\'e} J{\'e}gou, Julien Mairal,
  Piotr Bojanowski, and Armand Joulin.
\newblock Emerging properties in self-supervised vision transformers.
\newblock In {\em ICCV}, pages 9650--9660. IEEE, 2021.

\bibitem{chen2020simple}
Ting Chen, Simon Kornblith, Mohammad Norouzi, and Geoffrey Hinton.
\newblock A simple framework for contrastive learning of visual
  representations.
\newblock In {\em ICML}, pages 1597--1607. PMLR, 2020.

\bibitem{chen2022context}
Xiaokang Chen, Mingyu Ding, Xiaodi Wang, Ying Xin, Shentong Mo, Yunhao Wang,
  Shumin Han, Ping Luo, Gang Zeng, and Jingdong Wang.
\newblock Context autoencoder for self-supervised representation learning.
\newblock {\em arXiv preprint arXiv:2202.03026}, 2022.

\bibitem{chen2021mocov3}
Xinlei Chen*, Saining Xie*, and Kaiming He.
\newblock An empirical study of training self-supervised vision transformers.
\newblock {\em arXiv preprint arXiv:2104.02057}, 2021.

\bibitem{dalal2005histograms}
Navneet Dalal and Bill Triggs.
\newblock Histograms of oriented gradients for human detection.
\newblock In {\em CVPR}, volume~1, pages 886--893. IEEE, 2005.

\bibitem{deng2009imagenet}
Jia Deng, Wei Dong, Richard Socher, Li-Jia Li, Kai Li, and Li Fei-Fei.
\newblock Imagenet: A large-scale hierarchical image database.
\newblock In {\em CVPR}, pages 248--255. IEEE, 2009.

\bibitem{devlin2018bert}
Jacob Devlin, Ming-Wei Chang, Kenton Lee, and Kristina Toutanova.
\newblock Bert: Pre-training of deep bidirectional transformers for language
  understanding.
\newblock {\em arXiv preprint arXiv:1810.04805}, 2018.

\bibitem{dong2021peco}
Xiaoyi Dong, Jianmin Bao, Ting Zhang, Dongdong Chen, Weiming Zhang, Lu Yuan,
  Dong Chen, Fang Wen, and Nenghai Yu.
\newblock Peco: Perceptual codebook for bert pre-training of vision
  transformers.
\newblock {\em arXiv preprint arXiv:2111.12710}, 2021.

\bibitem{dosovitskiy2020image}
Alexey Dosovitskiy, Lucas Beyer, Alexander Kolesnikov, Dirk Weissenborn,
  Xiaohua Zhai, Thomas Unterthiner, Mostafa Dehghani, Matthias Minderer, Georg
  Heigold, Sylvain Gelly, et~al.
\newblock An image is worth 16x16 words: Transformers for image recognition at
  scale.
\newblock {\em arXiv preprint arXiv:2010.11929}, 2020.

\bibitem{fang2022corrupted}
Yuxin Fang, Li Dong, Hangbo Bao, Xinggang Wang, and Furu Wei.
\newblock Corrupted image modeling for self-supervised visual pre-training.
\newblock {\em arXiv preprint arXiv:2202.03382}, 2022.

\bibitem{feichtenhofer2019slowfast}
Christoph Feichtenhofer, Haoqi Fan, Jitendra Malik, and Kaiming He.
\newblock Slowfast networks for video recognition.
\newblock In {\em Proceedings of the IEEE/CVF international conference on
  computer vision}, pages 6202--6211, 2019.

\bibitem{he2022masked}
Kaiming He, Xinlei Chen, Saining Xie, Yanghao Li, Piotr Doll{\'a}r, and Ross
  Girshick.
\newblock Masked autoencoders are scalable vision learners.
\newblock In {\em CVPR}, pages 16000--16009. IEEE, 2022.

\bibitem{he2020momentum}
Kaiming He, Haoqi Fan, Yuxin Wu, Saining Xie, and Ross Girshick.
\newblock Momentum contrast for unsupervised visual representation learning.
\newblock In {\em CVPR}, pages 9729--9738. IEEE, 2020.

\bibitem{he2017mask}
Kaiming He, Georgia Gkioxari, Piotr Doll{\'a}r, and Ross Girshick.
\newblock Mask r-cnn.
\newblock In {\em ICCV}, pages 2961--2969. IEEE, 2017.

\bibitem{he2016deep}
Kaiming He, Xiangyu Zhang, Shaoqing Ren, and Jian Sun.
\newblock Deep residual learning for image recognition.
\newblock In {\em Proceedings of the IEEE conference on computer vision and
  pattern recognition}, pages 770--778, 2016.

\bibitem{imagenetr}
Dan Hendrycks, Steven Basart, Norman Mu, Saurav Kadavath, Frank Wang, Evan
  Dorundo, Rahul Desai, Tyler Zhu, Samyak Parajuli, Mike Guo, Dawn Song, Jacob
  Steinhardt, and Justin Gilmer.
\newblock The many faces of robustness: A critical analysis of
  out-of-distribution generalization.
\newblock {\em ICCV}, 2021.

\bibitem{imagenetc}
Dan Hendrycks and Thomas Dietterich.
\newblock Benchmarking neural network robustness to common corruptions and
  perturbations.
\newblock {\em ICLR}, 2019.

\bibitem{imageneta}
Dan Hendrycks, Kevin Zhao, Steven Basart, Jacob Steinhardt, and Dawn Song.
\newblock Natural adversarial examples.
\newblock {\em CVPR}, 2021.

\bibitem{hochreiter2001gradient}
Sepp Hochreiter, Yoshua Bengio, Paolo Frasconi, J{\"u}rgen Schmidhuber, et~al.
\newblock Gradient flow in recurrent nets: the difficulty of learning long-term
  dependencies, 2001.

\bibitem{hou2022milan}
Zejiang Hou, Fei Sun, Yen-Kuang Chen, Yuan Xie, and Sun-Yuan Kung.
\newblock Milan: Masked image pretraining on language assisted representation.
\newblock {\em arXiv preprint arXiv:2208.06049}, 2022.

\bibitem{ioffe2015batch}
Sergey Ioffe and Christian Szegedy.
\newblock Batch normalization: Accelerating deep network training by reducing
  internal covariate shift.
\newblock In {\em International conference on machine learning}, pages
  448--456. PMLR, 2015.

\bibitem{kay2017kinetics}
Will Kay, Joao Carreira, Karen Simonyan, Brian Zhang, Chloe Hillier, Sudheendra
  Vijayanarasimhan, Fabio Viola, Tim Green, Trevor Back, Paul Natsev, et~al.
\newblock The kinetics human action video dataset.
\newblock {\em arXiv preprint arXiv:1705.06950}, 2017.

\bibitem{kornblith2019similarity}
Simon Kornblith, Mohammad Norouzi, Honglak Lee, and Geoffrey Hinton.
\newblock Similarity of neural network representations revisited.
\newblock In {\em International Conference on Machine Learning}, pages
  3519--3529. PMLR, 2019.

\bibitem{lee2015deeply}
Chen-Yu Lee, Saining Xie, Patrick Gallagher, Zhengyou Zhang, and Zhuowen Tu.
\newblock Deeply-supervised nets.
\newblock In {\em Artificial intelligence and statistics}, pages 562--570.
  PMLR, 2015.

\bibitem{li2022semmae}
Gang Li, Heliang Zheng, Daqing Liu, Chaoyue Wang, Bing Su, and Changwen Zheng.
\newblock Semmae: Semantic-guided masking for learning masked autoencoders.
\newblock {\em arXiv preprint arXiv:2206.10207}, 2022.

\bibitem{li2022comprehensive}
Renjie Li, Xinyi Wang, Guan Huang, Wenli Yang, Kaining Zhang, Xiaotong Gu,
  Son~N Tran, Saurabh Garg, Jane Alty, and Quan Bai.
\newblock A comprehensive review on deep supervision: Theories and
  applications.
\newblock {\em arXiv preprint arXiv:2207.02376}, 2022.

\bibitem{lin2014microsoft}
Tsung-Yi Lin, Michael Maire, Serge Belongie, James Hays, Pietro Perona, Deva
  Ramanan, Piotr Doll{\'a}r, and C~Lawrence Zitnick.
\newblock Microsoft coco: Common objects in context.
\newblock In {\em ECCV}, pages 740--755. Springer, 2014.

\bibitem{liu2021video}
Ze Liu, Jia Ning, Yue Cao, Yixuan Wei, Zheng Zhang, Stephen Lin, and Han Hu.
\newblock Video swin transformer.
\newblock In {\em CVPR}, 2022.

\bibitem{tanet}
Zhaoyang Liu, Limin Wang, Wayne Wu, Chen Qian, and Tong Lu.
\newblock Tam: Temporal adaptive module for video recognition.
\newblock In {\em ICCV}, 2021.

\bibitem{loshchilov2018decoupled}
Ilya Loshchilov and Frank Hutter.
\newblock Decoupled weight decay regularization.
\newblock In {\em International Conference on Learning Representations}, 2018.

\bibitem{motionformer}
Mandela Patrick, Dylan Campbell, Yuki Asano, Ishan Misra, Florian Metze,
  Christoph Feichtenhofer, Andrea Vedaldi, and Joao~F. Henriques.
\newblock Keeping your eye on the ball: Trajectory attention in video
  transformers.
\newblock In {\em NeurIPS}, 2021.

\bibitem{peng2022beit}
Zhiliang Peng, Li Dong, Hangbo Bao, Qixiang Ye, and Furu Wei.
\newblock Beit v2: Masked image modeling with vector-quantized visual
  tokenizers.
\newblock {\em arXiv preprint arXiv:2208.06366}, 2022.

\bibitem{radford2021learning}
Alec Radford, Jong~Wook Kim, Chris Hallacy, Aditya Ramesh, Gabriel Goh,
  Sandhini Agarwal, Girish Sastry, Amanda Askell, Pamela Mishkin, Jack Clark,
  et~al.
\newblock Learning transferable visual models from natural language
  supervision.
\newblock In {\em International Conference on Machine Learning}, pages
  8748--8763. PMLR, 2021.

\bibitem{clip}
Alec Radford, Jong~Wook Kim, Chris Hallacy, A. Ramesh, Gabriel Goh, Sandhini
  Agarwal, Girish Sastry, Amanda Askell, Pamela Mishkin, Jack Clark, Gretchen
  Krueger, and Ilya Sutskever.
\newblock Learning transferable visual models from natural language
  supervision.
\newblock In {\em ICML}, 2021.

\bibitem{ramesh2021zero}
Aditya Ramesh, Mikhail Pavlov, Gabriel Goh, Scott Gray, Chelsea Voss, Alec
  Radford, Mark Chen, and Ilya Sutskever.
\newblock Zero-shot text-to-image generation.
\newblock In {\em ICML}, pages 8821--8831. PMLR, 2021.

\bibitem{ren2023tinymim}
Sucheng Ren, Fangyun Wei, Zheng Zhang, and Han Hu.
\newblock Tinymim: An empirical study of distilling mim pre-trained models.
\newblock 2023.

\bibitem{szegedy2015going}
Christian Szegedy, Wei Liu, Yangqing Jia, Pierre Sermanet, Scott Reed, Dragomir
  Anguelov, Dumitru Erhan, Vincent Vanhoucke, and Andrew Rabinovich.
\newblock Going deeper with convolutions.
\newblock In {\em Proceedings of the IEEE conference on computer vision and
  pattern recognition}, pages 1--9, 2015.

\bibitem{tao2022siamese}
Chenxin Tao, Xizhou Zhu, Gao Huang, Yu Qiao, Xiaogang Wang, and Jifeng Dai.
\newblock Siamese image modeling for self-supervised vision representation
  learning.
\newblock {\em arXiv preprint arXiv:2206.01204}, 2022.

\bibitem{tong2022videomae}
Zhan Tong, Yibing Song, Jue Wang, and Limin Wang.
\newblock Video{MAE}: Masked autoencoders are data-efficient learners for
  self-supervised video pre-training.
\newblock In {\em Advances in Neural Information Processing Systems}, 2022.

\bibitem{deit}
Hugo Touvron, Matthieu Cord, Matthijs Douze, Francisco Massa, Alexandre
  Sablayrolles, and Herve Jegou.
\newblock Training data-efficient image transformers \& distillation through
  attention.
\newblock In {\em International Conference on Machine Learning}, volume 139,
  pages 10347--10357, July 2021.

\bibitem{wang2015training}
Liwei Wang, Chen-Yu Lee, Zhuowen Tu, and Svetlana Lazebnik.
\newblock Training deeper convolutional networks with deep supervision.
\newblock {\em arXiv preprint arXiv:1505.02496}, 2015.

\bibitem{tdn}
Limin Wang, Zhan Tong, Bin Ji, and Gangshan Wu.
\newblock {TDN}: Temporal difference networks for efficient action recognition.
\newblock In {\em CVPR}, 2021.

\bibitem{nonlocal}
Xiaolong Wang, Ross Girshick, Abhinav Gupta, and Kaiming He.
\newblock Non-local neural networks.
\newblock In {\em CVPR}, 2018.

\bibitem{wang2021dense}
Xinlong Wang, Rufeng Zhang, Chunhua Shen, Tao Kong, and Lei Li.
\newblock Dense contrastive learning for self-supervised visual pre-training.
\newblock In {\em Proceedings of the IEEE/CVF Conference on Computer Vision and
  Pattern Recognition}, pages 3024--3033, 2021.

\bibitem{wei2022masked}
Chen Wei, Haoqi Fan, Saining Xie, Chao-Yuan Wu, Alan Yuille, and Christoph
  Feichtenhofer.
\newblock Masked feature prediction for self-supervised visual pre-training.
\newblock In {\em CVPR}, pages 14668--14678. IEEE, 2022.

\bibitem{wu2018unsupervised}
Zhirong Wu, Yuanjun Xiong, Stella~X Yu, and Dahua Lin.
\newblock Unsupervised feature learning via non-parametric instance
  discrimination.
\newblock In {\em Proceedings of the IEEE conference on computer vision and
  pattern recognition}, pages 3733--3742, 2018.

\bibitem{xiao2018unified}
Tete Xiao, Yingcheng Liu, Bolei Zhou, Yuning Jiang, and Jian Sun.
\newblock Unified perceptual parsing for scene understanding.
\newblock In {\em ECCV}, pages 418--434. Springer, 2018.

\bibitem{xie2015holistically}
Saining Xie and Zhuowen Tu.
\newblock Holistically-nested edge detection.
\newblock In {\em Proceedings of the IEEE international conference on computer
  vision}, pages 1395--1403, 2015.

\bibitem{xie2022revealing}
Zhenda Xie, Zigang Geng, Jingcheng Hu, Zheng Zhang, Han Hu, and Yue Cao.
\newblock Revealing the dark secrets of masked image modeling.
\newblock {\em arXiv preprint arXiv:2205.13543}, 2022.

\bibitem{xie2021propagate}
Zhenda Xie, Yutong Lin, Zheng Zhang, Yue Cao, Stephen Lin, and Han Hu.
\newblock Propagate yourself: Exploring pixel-level consistency for
  unsupervised visual representation learning.
\newblock In {\em Proceedings of the IEEE/CVF Conference on Computer Vision and
  Pattern Recognition}, pages 16684--16693, 2021.

\bibitem{xie2022simmim}
Zhenda Xie, Zheng Zhang, Yue Cao, Yutong Lin, Jianmin Bao, Zhuliang Yao, Qi
  Dai, and Han Hu.
\newblock Simmim: A simple framework for masked image modeling.
\newblock In {\em CVPR}, pages 9653--9663. IEEE, 2022.

\bibitem{xie2022data}
Zhenda Xie, Zheng Zhang, Yue Cao, Yutong Lin, Yixuan Wei, Qi Dai, and Han Hu.
\newblock On data scaling in masked image modeling.
\newblock {\em arXiv preprint arXiv:2206.04664}, 2022.

\bibitem{zhang2022contrastive}
Linfeng Zhang, Xin Chen, Junbo Zhang, Runpei Dong, and Kaisheng Ma.
\newblock Contrastive deep supervision.
\newblock In {\em Computer Vision--ECCV 2022: 17th European Conference, Tel
  Aviv, Israel, October 23--27, 2022, Proceedings, Part XXVI}, pages 1--19.
  Springer, 2022.

\bibitem{ade20k}
Bolei Zhou, Hang Zhao, Xavier Puig, Tete Xiao, Sanja Fidler, Adela Barriuso,
  and Antonio Torralba.
\newblock Semantic understanding of scenes through the ade20k dataset.
\newblock {\em IJCV}, 127(3):302--321, 2019.

\bibitem{zhou2021ibot}
Jinghao Zhou, Chen Wei, Huiyu Wang, Wei Shen, Cihang Xie, Alan Yuille, and Tao
  Kong.
\newblock ibot: Image bert pre-training with online tokenizer.
\newblock {\em ICLR}, 2022.

\end{thebibliography}
}
\appendix
In this supplementary material, we first provide further implementation details (\cref{sec:implementation}).
We then describe additional ablation studies (\cref{sec:ablations}).

\section{More Implementation Details}
\label{sec:implementation}

\noindent\textbf{Pretraining.} We follow the same training recipe of MAE~\cite{he2022masked} and use a batch size of 4096 and a learning rate of $lr = base\_lr \times \text{BatchSize}/256$, where $base\_lr=$1.5e-4. We adopt a cosine decay schedule with a warm-up for 40 epochs.  We use AdamW~\cite{loshchilov2018decoupled} with a weight decay of 0.05 as the optimizer. For data augmentations, we use random resized cropping and random horizontal flipping. The mask ratio is set to 75\% by default.

\noindent\textbf{Finetuning on ImageNet-1K.} ImageNet-1K~\cite{deng2009imagenet} contains 1.28M images for training and 50K images for validation. We follow the same recipe of MAE~\cite{he2022masked} on ImageNet-1K except for the learning rate and layer decay. We adopt AdamW as the optimizer and finetune the pre-trained model for 100 epochs. The learning rate is set to 2e-4 for DeepMIM on MAE-Pixel~\cite{he2022masked}/HOG~\cite{dalal2005histograms}/DINO~\cite{caron2021emerging}/Codebook~\cite{dong2021peco} and 4e-4 for DeepMIM on MAE-CLIP. We adopt a cosine decay schedule with a warm-up for 5 epochs. The layer-wise learning rate decay is set to 0.65 for 300-epoch pre-trained model and 0.6 for 800-epoch pre-trained model. The batch size is set to 1024. For data augmentation and regularization, we adopt drop path with a drop path rate of 0.1, RandAugment, label smoothing of 0.1, Mixup of 0.8, CutMix of 1.0.

\noindent\textbf{Finetuning on COCO Objection Detection and Instance Segmentation.} COCO~\cite{lin2014microsoft} contains 118K images for training and 5K images for validation. Following\cite{he2022masked}, we adopt DeepMIM pre-trained ViT as the backbone and Mask-RCNN~\cite{he2017mask} as the framework. For all DeepMIM pre-trained models, we use AdamW as the optimizer with a weight decay of 0.1. The learning rate, the layer-wise learning rate decay and the batch size is set to 1e-4, 0.75 and 16, respectively. The input resolution is 1024$\times$1024. We do not use multi-scale testing.

\noindent\textbf{Finetuning on ADK20K Semantic Segmentation.} ADE20K~\cite{ade20k} is a widely used dataset for semantic segmentation with 150 classes and 20K images for training and 2K images for validation. We use DeepMIM pre-trained ViT as the backbone and UperNet~\cite{xiao2018unified} as the segmentation model. We use AdamW as the optimizer with a weight decay of 0.05. We set drop path rate to 0.1, batch size to 16. For DeepMIM on MAE, the learning rate is set to 1e-4, the layer-wise learning rate decay is set to 0.65. For DeepMIM on MAE-CLIP, the learning rate is set to 5e-5, the layer-wise learning rate decay is set to 0.9. The input resolution is 512$\times$512. We do not use multi-scale testing.

\noindent\textbf{Pre-training and Finetuning on Kinetics-400 Video Classification.} DeepMIM on VideoMAE~\cite{tong2022videomae} is pre-trained for 800 epochs and finetuned for 75 epochs on Kinetics-400 video classification dataset. The optimizer, data augmentation and hyper-parameters are the same as when pre-training on ImageNet. We use a batch size of 1024 during pre-training. During finetuning, we use AdamW with a weight decay of 0.05 as the optimizer. The learning rate is set to 1e-3. We adopt a cosine decay schedule with a warm-up for 5 epochs. The layer-wise learning rate decay is set to 0.75. For data augmentation and regularization, we adopt drop path with a drop path rate of 0.1, RandAugment, label smoothing of 0.1, Mixup of 0.8, CutMix of 1.0. We follow SlowFast~\cite{feichtenhofer2019slowfast} and take 5 clips $\times$ 3 crops for inference.

\section{More Ablation Studies}
\label{sec:ablations}
\begin{table}[t]
	\centering
	\begin{tabular}{lcl}
		\toprule
		Method    & Masking Strategy & Top-1 Acc. \\
		\midrule
		MAE  & Random~\cite{he2022masked} & 82.6 \\
		DeepMIM  & Random~\cite{he2022masked} & 83.6 (+1.0) \\
		\midrule
		MAE  & Semantic-guided~\cite{li2022semmae} & 82.8 \\
		DeepMIM  & Semantic-guided~\cite{li2022semmae} & 83.8 (+1.0) \\
		\midrule
		MAE  & Semantic-aware~\cite{hou2022milan} & 82.9 \\
		DeepMIM  & Semantic-aware~\cite{hou2022milan}  & 83.8 (+0.9) \\
		\bottomrule
	\end{tabular}
		\caption{DeepMIM is compatible with different masking strategies. The results are evaluated on ImageNet-1K finetuning.}
	\label{tab:masking}
\end{table}
\noindent\textbf{DeepMIM is Compatible with Different Masking Strategies.}
DeepMIM is compatible not only with different reconstruction targets at output level, but also with different masking strategies at input level. As shown in Table~\ref{tab:masking}, DeepMIM consistently improves various MIM models with different masking strategies including random masking~\cite{he2022masked}, semantic-guided masking~\cite{li2022semmae} and semantic-aware masking~\cite{hou2022milan}.

\end{document}